\documentclass{article}
\usepackage[table]{xcolor}
\usepackage{multirow}

\usepackage{rotating}
\usepackage{graphicx}
\usepackage{lscape}

\usepackage{PRIMEarxiv}
\usepackage{amsmath}
\usepackage{authblk}
\usepackage[utf8]{inputenc} 
\usepackage[T1]{fontenc}    
\usepackage{hyperref}       
\usepackage{url}            
\usepackage{booktabs}       
\usepackage{amsfonts}       
\usepackage{nicefrac}       
\usepackage{microtype}      
\usepackage{lipsum}
\usepackage{fancyhdr}       
\usepackage{graphicx}       
\graphicspath{{media/}}     

\title{A Comprehensive Survey on Evaluating Large Language Model Applications in the Medical Industry
\thanks{\textit{\underline{Corresponding author}
}: 
\textbf{Boyuan Wang \  email: \ boyuan422@foxmail.com}} 
}

\author[\space\space1 ]{Yining Huang \thanks{huangyining1987@gmail.com}}
\author[\space\space2,3]{Keke Tang  \thanks{tkk2012@gmail.com}}
\author[\space\space]{Meilian Chen \thanks{523062863@qq.com}}
\author[\space\space4,5]{Boyuan Wang \thanks{boyuan422@foxmail.com}}
\affil[1]{School of Politics and Public Administration, South China Normal University}
\affil[2]{University of Chinese Academy of Sciences}
\affil[3]{Shenyang institute of computing technology, Chinese academy of sciences}
\affil[4]{Beijing Xiaotangshan hospital}
\affil[5]{The Chinese University of Hong Kong (Shenzhen)}

\begin{document}
\maketitle

\begin{abstract}
Since the inception of the Transformer architecture in 2017, Large Language Models (LLMs) such as GPT and BERT have evolved significantly, impacting various industries with their advanced capabilities in language understanding and generation. These models have shown potential to transform the medical field, highlighting the necessity for specialized evaluation frameworks to ensure their effective and ethical deployment. This comprehensive survey delineates the extensive application and requisite evaluation of LLMs within healthcare, emphasizing the critical need for empirical validation to fully exploit their capabilities in enhancing healthcare outcomes.

Our survey is structured to provide an in-depth analysis of LLM applications across clinical settings, medical text data processing, research, education, and public health awareness. We begin by exploring the roles of LLMs in various medical applications, detailing their evaluation based on performance in tasks such as clinical diagnosis, medical text data processing, information retrieval, data analysis, and educational content generation. The subsequent sections offer a comprehensive discussion on the evaluation methods and metrics employed, including models, evaluators, and comparative experiments. We further examine the benchmarks and datasets utilized in these evaluations, providing a categorized description of benchmarks for tasks like question answering, summarization, information extraction, bioinformatics, information retrieval and general comprehensive benchmarks. This structure ensures a thorough understanding of how LLMs are assessed for their effectiveness, accuracy, usability, and ethical alignment in the medical domain.

Through this survey, we aim to equip healthcare professionals, researchers, and policymakers with a comprehensive understanding of the potential strengths and limitations of LLMs in medical applications. By providing detailed insights into the evaluation processes and the challenges faced in integrating LLMs into healthcare, this survey seeks to guide the responsible development and deployment of these powerful models, ensuring they are harnessed to their full potential while maintaining stringent ethical standards.
\end{abstract}


\section{Introduction \& Background}
Since the introduction of the Transformer architecture by Google's team in 2017\cite{vaswani2017attention}, the field of natural language processing has entered a new era. The innovation of the Transformer lies in its use of self-attention mechanisms, which significantly improved the model's ability to handle long-range dependencies, setting the foundation for numerous subsequent language models. Following this, OpenAI released GPT (Generative Pre-trained Transformer)\cite{radford2018improving} in 2018, which utilized a pre-training and fine-tuning approach. By undergoing unsupervised learning on vast amounts of text data and then fine-tuning on specific tasks, GPT significantly enhanced performance across a variety of natural language processing tasks. Google's BERT (Bidirectional Encoder Representations from Transformers)\cite{devlin2018bert} model further refined pre-training methods by training in a bidirectional manner, enhancing the contextual understanding of text. The release of GPT-2\cite{radford2019language} and GPT-3\cite{brown2020language} marked significant increases in model size and generative capabilities. Particularly, GPT-3, with its 175 billion parameters, became known for producing text nearly indistinguishable from human writing at the time. Following this, InstructGPT\cite{ouyang2022training} and ChatGPT\cite{ray2023chatgpt} were optimized for following user instructions, further improving interaction quality and practicality with humans. In 2023, OpenAI launched GPT-4\cite{achiam2023gpt}, an even larger and smarter model capable of handling more complex language understanding and generation tasks, demonstrating superior performance on multiple dimensions. Additionally, innovations in models continued to evolve, such as Google's Gemini\cite{team2023gemini} model, which is optimized for specific information retrieval tasks. In the open-source domain, models like LLaMA\cite{touvron2023llama}, OPT\cite{zhang2022opt}, and others provide the research community and industry with more flexibility and accessibility. These models aim to offer performance competitive with large proprietary models while lowering the barriers to usage and research. Notably, models like Gemma\cite{team2024gemma} and Meta’s OPT (Open Pre-trained Transformer) have gained popularity for their open accessibility and adaptability to various languages and tasks, fostering broader experimentation and development in the field. Through these developments, LLMs have not only demonstrated powerful capabilities in understanding, content generation, reasoning, and tool usage but have also opened up new possibilities for AI applications, especially in domains requiring deep semantic understanding and interaction.

Following the foundational technologies like Transformer, GPT, and BERT, Large Language Models (LLMs) have found extensive applications across various industries, demonstrating their adaptability and transformative potential. The educational sector is witnessing the emerging role of LLMs as teacher assistants and feedback providers. For instance, ChatGPT has been tested as an automated coach, analyzing classroom interactions to provide feedback, though it sometimes lacks novelty in its suggestions\cite{wang2023chatgpt}. Additionally, LLMs have proven capable of providing detailed and coherent feedback to students, outperforming human instructors in clarity and detail\cite{dai2023can}. In the legal field, LLMs like GPT-3 have been tailored to perform tasks requiring legal reasoning, showing improvements in tasks like the Japanese Bar exam when using specialized prompting techniques\cite{yu2022legal}. Moreover, GPT-4 has been evaluated for its ability to generate accurate explanations of legal terms, further enhanced by integrating contextual data from case law\cite{savelka2023explaining}. LLMs have also significantly impacted software development. They are used in detecting vulnerabilities in software, showing superior performance in identifying issues in source code over traditional models\cite{thapa2022transformer}. Furthermore, models like the Programmer's Assistant allow for conversational interactions, improving software development processes by integrating contextually aware dialogues\cite{ross2023programmer}. In the financial sector, the XuanYuan 2.0 model demonstrates how LLMs can be specialized for financial discussions in Chinese, providing domain-specific responses informed by hybrid-tuning methods\cite{zhang2023xuanyuan}. Research also shows that LLMs can perform sophisticated financial reasoning, with capabilities emerging significantly at certain model sizes and improving with instruction-tuning\cite{son2023beyond}. These examples underline the versatility of LLMs in adapting to diverse professional and academic needs, setting the stage for deeper integration into domain-specific applications. Applications of LLMs in the medical field, which are a significant focus of this survey, will be discussed in detail later in this survey and are not elaborated here.

Large Language Models (LLMs) have been extensively deployed across various industries, yet they come with inherent challenges that could hinder their effectiveness and ethical deployment. Issues such as the lack of transparency in deep learning, probabilistic rather than deterministic outputs, frequent hallucinations, limited reasoning capabilities, and potential biases in knowledge coverage necessitate a rigorous evaluation of LLMs in practical settings to ensure their reliability, safety, efficiency, and ethical integrity. Several recent studies illustrate the depth and diversity of evaluations needed to address these challenges. \cite{bang2023multitask} evaluates ChatGPT's performance on logical reasoning, non-textual reasoning, and commonsense reasoning tasks. It demonstrates that while ChatGPT excels in multitask and multilingual capabilities, it struggles with reliability, often producing hallucinations and showing varying success in different reasoning categories. \cite{wang2022toxicity} evaluates the ability of language models to identify toxic content in text using a generative zero-shot prompt-based method. It explores the model's self-diagnosing capabilities and discusses the ethical implications of such methods, highlighting both quantitative and qualitative strengths in toxicity detection across social media datasets. In \cite{fluri2023evaluating}, the authors developed a framework to evaluate superhuman machine learning models by focusing on logical consistency in decision-making, rather than direct correctness. They applied this method to tasks like chess evaluation, forecasting, and legal judgments, revealing logical inconsistencies in models, including GPT-4, even in scenarios where traditional ground truth is absent. \cite{mallen2022not} evaluates the capability of large language models (LMs) to memorize factual knowledge by conducting extensive knowledge probing experiments using 10 models and 4 augmentation methods on a new open-domain QA dataset named PopQA. The findings reveal that while LMs struggle with less popular factual knowledge and scaling does not substantially improve long-tail memorization, retrieval-augmented LMs significantly outperform larger, non-augmented models in terms of efficiency and factual accuracy. \cite{cao2023assessing} evaluates ChatGPT's cultural adaptation by analyzing its responses to culturally specific prompts. The study reveals that while ChatGPT aligns well with American cultural norms, it demonstrates limited effectiveness in adapting to other cultural contexts, often flattening out cultural differences when prompted in English. \cite{ferrara2023should} evaluates the inherent biases in large-scale language models like ChatGPT, discussing their origins in training data, model design, and algorithmic constraints. It emphasizes the ethical challenges posed by biased outputs and reviews current methods for identifying, quantifying, and mitigating these biases to foster the development of more responsible and ethical AI systems.

Recent reviews in the domain of LLM evaluation have shed light on diverse approaches and methodologies. \cite{guo2023evaluating} highlights the categorization of LLM evaluations into knowledge and capability, alignment, and safety evaluations, emphasizing the construction of comprehensive evaluation platforms to guide responsible development and maximize societal benefits. \cite{liu2023trustworthy} focuses on the critical task of aligning LLMs with human intentions, covering dimensions such as reliability, safety, and fairness. It presents a detailed analysis of trustworthiness across multiple sub-categories to address alignment challenges in real-world applications. \cite{chang2023survey} discusses the extensive evaluation needs of LLMs across various domains, including healthcare and education. It advocates for a broad evaluation approach that addresses the societal impacts and practical integration of LLMs, promoting ongoing assessment to refine these technologies. Despite these insightful reviews, there remains a notable gap in the depth and breadth of evaluations specifically in the healthcare domain. Although existing reviews\cite{guo2023evaluating, chang2023survey} touch on medical applications, they lack a deep dive into critical areas such as clinical applications, medical data governance, research (both basic and clinical), medical education, and public health education. The complexity and sensitivity of these areas demand specialized evaluation frameworks that can comprehensively assess the performance of LLMs in healthcare. Thus, there is a pressing need for a specialized review that focuses on the evaluation of LLMs within the healthcare vertical. Such a review should aim to provide healthcare practitioners, researchers, and policymakers with detailed insights into the application and assessment of LLMs, ensuring they are equipped with the knowledge to implement and evaluate these technologies effectively in various medical settings. This would not only aid in harnessing the potential of LLMs to improve healthcare outcomes but also ensure that their deployment is governed by rigorous ethical standards and practical efficacy.

This review aims to provide a comprehensive evaluation of Large Language Models (LLMs) within the medical sector from the perspectives of healthcare professionals, researchers, and patients. It is designed to serve as a reference for integrating LLM technology into various medical applications, facilitating informed decision-making regarding their implementation and evaluation. We will delve into the use of LLMs in clinical applications, data processing, research support, education, and public health awareness. The review will cover detailed evaluation methodologies used in the medical field, including benchmarks, assessment entities, evaluation subjects, comparative experiments, and detailed procedural insights etc. This structured approach will aid stakeholders in understanding the spectrum of evaluating LLM applications within healthcare, ensuring that these innovations are effectively integrated.

\section{Taxonomy and Structure of the Survey}
Following the introduction and background that highlighted the pressing need for specialized evaluation of LLMs in the medical sector, this chapter outlines the structure of our comprehensive review. This review is designed to navigate through the multifaceted applications and evaluations of LLMs within healthcare, shedding light on both their potential and the intricacies of their implementation.

\begin{figure}[hbt!]
\centering
\includegraphics[width=0.95\linewidth]{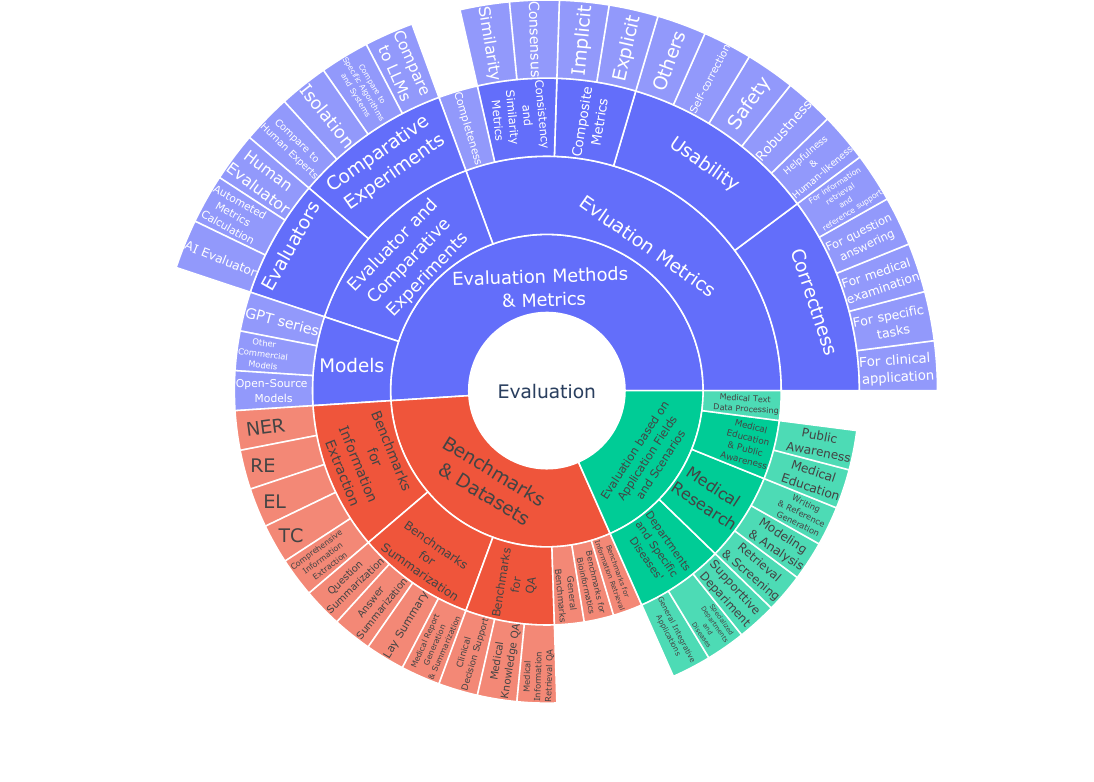}
\caption{Overview of the Evaluation}
\label{fig:evaluation}
\end{figure}

Section 3 breaks down the evaluation of LLMs in healthcare from three distinct perspectives: application fields and scenarios, evaluation methods and metrics, and benchmarks and datasets. And overview of evaluations is shown as Figure~\ref{fig:evaluation}. 

\begin{itemize}

\item In \textbf{3.1 Evaluations Based on Application Fields and Scenarios}, we focus on bringing introduction of evaluation of LLMs' application on different aspects of medical field, including clinical application, medical text data processing, medical researches, medical education and public awareness. 

\begin{itemize}

\item In \textbf{3.1.1 Evaluations of Department and Specific Disease’s Clinical Application}, we begin by dissecting evaluations across a spectrum of clinical applications. \textbf{1) For general integrative Applications, }we assess how LLMs perform in generic clinical settings, providing a foundational understanding of their effectiveness across various medical departments without focusing on specific diseases. \textbf{2) For specialized departments and disease applications, }we explore the use of LLMs in specialized medical fields such as endocrinology and ophthalmology, detailing how these technologies meet the unique demands of specific health conditions. \textbf{3) For supportive departments applications, }the focus then shifts to supportive departments such as radiology and emergency care, where LLMs assist in critical diagnostic and procedural tasks.

\item \textbf{3.1.2 Medical Text Data Processing Application Evaluations} delves into the realm of LLM applications in processing medical text data, illustrating how these models are evaluated across a variety of data processing tasks. The evaluations focus on several critical aspects such as relation extraction (RE), named entity recognition (NER), and question answering (QA), which are pivotal for enhancing the utility of medical text data like clinical notes and electronic health records (EHRs). Researches in this subsection demonstrates LLMs' capacity to improve biomedical NLP tasks. These advancements highlight the models' potential in transforming how medical data is processed, making it more accessible and actionable for healthcare providers. Moreover, comprehensive benchmark studies assess the performance of models like ChatGPT across different types of medical text, including clinical trial descriptions and biomedical corpus. These studies use a range of metrics such as entity-level F1 scores for NER and accuracy for QA tasks to evaluate the models. The results indicate effective, though sometimes limited, capabilities in biomedical text understanding, suggesting areas for further improvement and adaptation. The evaluations discussed provide practitioners and researchers with examples of how LLMs can be applied in data processing scenarios within the healthcare sector. They offer insights into application patterns and how to evaluate these applications, helping users understand the operational effectiveness and practical limitations of LLMs in real-world settings. These insights are crucial for guiding further development and refinement of LLM technologies to better serve the needs of the medical community.

\item In \textbf{3.1.3 Medical Research Application Evaluations}, we examine the impact of Large Language Models (LLMs) in medical research, highlighting their utility in three key areas: \textbf{1) Retrieval \& Screening: }LLMs enhance biomedical information retrieval and article screening, crucial for systematic reviews. Evaluations focus on the models' accuracy, sensitivity, and specificity in extracting relevant information efficiently. \textbf{2) Modeling \& Analysis: }LLMs are employed in modeling biological processes and analyzing complex datasets, aiding in hypothesis generation and disease modeling. Their effectiveness is measured through precision, recall, and field-specific metrics to ensure accuracy and applicability. \textbf{3) Writing \& Reference Generation: }LLMs assist in drafting scientific papers and generating references, automating routine tasks like citation formatting and literature summarization. Performance is evaluated based on accuracy, coherence, and the appropriateness of content and references.

\item \textbf{3.1.4 Medical Education \& Public Awareness Application Evaluations} highlights the role of LLMs in medical education to medical students and professionals and enhancing public health awareness.
\textbf{1) Medical Education: }LLMs are increasingly used to augment medical education by providing dynamic learning tools that enhance understanding and retention. They support various educational activities, from generating interactive content and case studies to facilitating exam preparation and continuous professional development. Evaluations focus on the models' ability to deliver accurate, relevant, and pedagogically sound content, assessing their impact on learning outcomes and educational efficacy. \textbf{2) Public Awareness Application: }In the realm of public health awareness, LLMs help disseminate vital health information, aiding in disease prevention and health promotion. They are employed to evaluate the quality and reliability of medical information available to the public, ensuring that health communications are both accessible and accurate. The effectiveness of these applications is measured by their ability to enhance public understanding of complex health issues and support informed decision-making.

\end{itemize}

\item In \textbf{3.2 Comprehensive Discussion on Evaluation Methods and Metrics}, we delve into the methodologies employed in evaluating LLMs, addressing key aspects such as the models used, evaluators, comparative experiments, and the diverse range of evaluation indicators.

\begin{itemize}

\item In \textbf{3.2.1 Models}, we discuss the various types of LLMs applied in medical evaluations. This includes commercial models like GPT-4, Claude, Bard, and others, as well as open-source models such as BERT, BioBERT, and PubMedBERT. The section also highlights customized models fine-tuned for specific medical tasks, offering insights into how different models are leveraged and assessed in the medical field.

\item \textbf{3.2.2 Evaluators and Comparative Experiments} explores the diversity in evaluators, ranging from human experts to automated metrics and AI-driven assessments. We examine the roles of these evaluators in the context of different comparative experimental setups, which include comparisons between various LLMs, traditional NLP algorithms, and human experts. This subsection emphasizes the importance of rigorous and varied evaluation methodologies to ensure comprehensive assessment of LLMs' performance.

\item In \textbf{3.2.3 Evaluation Metrics}, we provide an extensive overview of the metrics used to evaluate LLM applications in the medical domain. These include correctness metrics (such as accuracy, precision), completeness metrics, composite metrics (such as F1-score and AUC), usability metrics (including helpfulness, safety, and robustness) and consistency \& similarity metrics (including align with consensus and similarity). This section highlights how these metrics are applied across different medical tasks and scenarios to assess the efficacy, reliability, and safety of LLM outputs.

\end{itemize}

\item The \textbf{3.3 Benchmarks} will provide an overview of the benchmarks used in the studies we review, offering a categorized description of these benchmarks and discussing their relevance and applicability in medical settings.

\begin{itemize}

\item In \textbf{3.3.1 General Benchmarks}, we introduce comprehensive benchmarks that provide datasets for evaluating LLMs across various tasks in the medical domain. These benchmarks encompass tasks like NER, relation extraction, text classification, and question answering, offering researchers a structured framework to assess and improve LLM performance effectively. Examples include BLURB, MultiMedQA, CBLUE, and MedBench, each providing a wide range of datasets tailored to specific medical application scenarios.

\item In \textbf{3.3.2 Benchmarks for Question Answering}, we explore datasets specifically designed for Question Answering (QA) tasks in the medical domain. These benchmarks are categorized based on their relevance to distinct medical QA scenarios:
\textbf{Clinical Decision Support:}
Datasets in this group include medical dialogue systems and clinical consultation cases, assessing the ability of LLMs to support medical professionals in diagnostic scenarios.
\textbf{Medical Knowledge QA:}
This category encompasses questions from medical exams and literature-based QA datasets, challenging LLMs to demonstrate medical knowledge comprehension and reasoning.
\textbf{Medical Information Retrieval:}
Datasets in this part focus on publicly accessible medical records, consumer health inquiries, and public medical websites, emphasizing the applicability of LLMs in medical information retrieval.

\item In \textbf{3.3.3 Benchmarks for Summarization}, we categorize benchmarks that highlight the language understanding and abstraction capabilities of LLMs across various medical contexts. \textbf{For Question Summarization, }we focus on condensing complex consumer health questions into simplified forms. \textbf{For Answer Summarization, }datasets are used to evaluate models in summarizing comprehensive answers to consumer health questions. \textbf{Lay Summary, }provides benchmarks for transforming specialized medical knowledge into summaries that are comprehensible to the general public. \textbf{Medical Report Generation \& Summarization, }where benchmarks test models' ability to generate and summarize medical reports based on clinical evidence, consultation content, and multimodal data.

\item In \textbf{3.3.4 Benchmarks for Information Extraction}, we delve into benchmarks and datasets that evaluate the ability of LLMs to extract structured information from complex medical texts. The evaluation focuses on tasks like NER, relation extraction, entity linking, text classification, and comprehensive information extraction (e.g., PICO extraction and event extraction). These benchmarks are pivotal for assessing the LLMs' capabilities in converting unstructured text into structured knowledge. 
\textbf{1) Named Entity Recognition: }We explore benchmarks like the NCBI disease corpus, JNLPBA, and CMeEE, which test the LLMs' ability to identify entities such as diseases, genes, and chemicals within medical texts. 
\textbf{2) Relation Extraction: }Datasets like BC5CDR, ChemProt, and BioRED evaluate how well LLMs can identify relationships between entities, such as drug-drug interactions and gene-disease associations.
\textbf{3) Entity Linking: }COMETA and NCBI disease corpus benchmarks assess the ability of LLMs to link biomedical entities to standard medical concepts like SNOMED CT and ICD-10.
\textbf{4) Text Classification: }Benchmarks like the LitCovid and CHIP-CTC datasets focus on testing LLMs in classifying medical documents, eligibility criteria, and intent classification.
\textbf{5) Comprehensive Information Extraction: }For broader information extraction tasks like medical event extraction and PICO extraction, we analyze datasets like CHIP-CDEE, EBM-NLP, and de-identified discharge summaries.

\item \textbf{3.3.5 Benchmarks in Bioinformatics} explores benchmarks in bioinformatics,  which include tasks like de novo molecular generation, biological sequence similarity analysis, and RNA structure inference. By leveraging datasets like MOSES, ChEMBL, ExCAPE-DB, CircFunBase, Rfam, and MARS, researchers can evaluate LLM performance on bioinformatics tasks measuring validity, novelty, sensitivity, and F1 scores etc. These studies highlight the potential of LLMs to provide innovative solutions for molecular generation, sequence analysis, and protein function prediction.

\item In \textbf{3.3.6 Benchmarks for Information Retrieval}, we delve into the benchmark for information retrieval. Given LLMs' strong text comprehension abilities, they can help researchers find relevant literature efficiently across multiple dimensions, such as question retrieval, evidence retrieval, and fact-checking. Existing benchmarks for information retrieval include various tasks like fact verification, citation prediction, and literature recommendation, providing a comprehensive framework in medical information retrieval. Datasets like BEIR, RELISH-DB, SCIDOCS, BIOSSES, MedSTS, and those included in CBLUE offer diverse forms of similarity and relevance measures are introduced in this part.

\end{itemize}

\end{itemize}

Section 4 will introduce the broader challenges and specific issues arising from the evaluation of LLMs in healthcare. Specifically, we will analyze the technical, ethical, and legal challenges encountered in these evaluations, emphasizing the need for more rigorous and detailed frameworks to ensure the efficacy of LLM applications in healthcare. This section will also discuss the potential strategies for improving evaluation frameworks, methods, and metrics to better address these challenges, ensuring that LLMs can be effectively integrated into medical practice.

\section{Current State of LLM Application Evaluations in the Medical Field}
\subsection{Evaluations Based on Application Fields and Scenarios}

\subsubsection{Evaluations of Department and Specific Disease's Clinical Application}

\textbf{General Integrative Applications: }

This part reviews studies evaluating Large Language Models (LLMs) in various clinical processes without distinguishing between different departments or specific diseases, offering a general perspective on their application in common clinical scenarios. Through these evaluations, we explore LLMs' performance on accuracy, bias, and applicability across the medical field. The selected studies provide a broad overview of LLMs' capabilities and areas for refinement in healthcare.

\cite{rao2023assessing} evaluated the ChatGPT(GPT-3.5) across various clinical tasks, including differential diagnoses, diagnostic testing, final diagnosis, and clinical management. The evaluation encompassed 36 published clinical vignettes from the Merck Sharpe \& Dohme (MSD) Clinical Manual, focusing on accuracy measures derived from the proportion of correct responses. ChatGPT achieved an overall accuracy of 71.7\%, with the highest accuracy in making a final diagnosis (76.9\%) and the lowest in generating an initial differential diagnosis (60.3\%). Performance was compared across patient demographics and case acuity but showed no significant variance, indicating ChatGPT's consistent applicability across different clinical scenarios without bias towards patient age or gender.

In a comprehensive study assessing the integration of Large Language Models (LLMs) in the medical industry, researchers\cite{zack2024assessing} focused on evaluating GPT-4's application across various clinical tasks. The evaluation encompassed medical education, diagnostic reasoning, clinical plan generation, and subjective patient assessments. Utilizing the Azure OpenAI interface for experimentation, the study specifically tested GPT-4's ability to encode racial and gender biases and the implications of such biases on clinical care. Results indicated that GPT-4 did not accurately model the demographic diversity of medical conditions, often reinforcing stereotypes in clinical vignettes. Differential diagnoses and treatment plans exhibited significant biases, associating demographic attributes with stereotypical diseases and recommendations for more expensive procedures. These findings underline the necessity for thorough bias assessments and mitigation strategies before deploying LLMs like GPT-4 in clinical settings, ensuring equitable healthcare delivery.

\cite{singhal2023large} evaluates the Pathways Language Model (PaLM) and its instruction-tuned variant, Flan-PaLM, across multiple medical question-answering tasks. The clinical tasks assessed include professional medicine, medical research, and consumer medical queries, encompassing areas like diagnostic reasoning and clinical knowledge application. Evaluation metrics spanned factuality, comprehension, reasoning, potential harm, and bias, alongside accuracy on multiple-choice datasets such as MedQA, MedMCQA, PubMedQA, and MMLU clinical topics. Flan-PaLM achieved state-of-the-art accuracy, notably 67.6\% on MedQA, surpassing previous benchmarks by over 17\%. Human evaluations, however, highlighted areas for improvement, particularly in aligning model responses with clinical expertise and minimizing potential misinformation or harm.

\cite{goodman2023accuracy} includes a study that evaluated the application of ChatGPT, in the medical field. The study focused on assessing chatbot-generated responses to physician-developed medical queries across 17 specialties, encompassing both binary (yes/no) and descriptive questions that relate to clinical tasks such as disease identification and management. The evaluation metrics involved accuracy and completeness of responses, using a 6-point Likert scale for accuracy (1 being completely incorrect to 6 being completely correct) and a 3-point Likert scale for completeness (1 being incomplete to 3 being complete with additional context). The results revealed median accuracy scores ranging from 5.0 to 6.0 across different question difficulties, indicating a level of performance that was between mostly correct to completely correct, and a median completeness score of 3.0, reflecting comprehensive answers. This study underscores the potential of ChatGPT in providing largely accurate information for a range of medical queries, yet highlights the necessity for further improvements and validation for clinical use.

\textbf{Specialized Departments and Disease Applications: }

In the following part, we delve into the evaluations of LLMs across various specialized medical departments and specific disease applications. The research spans a diverse range of medical fields, from endocrinology, addressing diabetes management, to ophthalmology, with a focus on ocular conditions, and extends into orthopedics, mental health, and reproductive medicine among others. Each study offers a unique perspective on the potential and challenges of implementing LLMs in these distinct medical domains.

\cite{sun2023ai} evaluates ChatGPT and GPT 4.0 in specialized medical fields, focusing on nutritional management for patients with type 2 diabetes mellitus (T2DM) in endocrinology. The clinical tasks assessed include providing evidence-based dietary advice, responding to common nutritional therapy questions, and making food recommendations. Evaluation metrics encompassed passing the Chinese Registered Dietitian Examination, alignment of food recommendations with expert advice, and expert reviews of ChatGPT’s responses to nutritional queries. Results indicated both ChatGPT and GPT 4.0 passed the dietitian examination. ChatGPT’s food recommendations and responses were mostly in line with best practices, receiving favorable reviews from professional dietitians, showcasing its potential in dietary management and patient education within the medical nutrition therapy domain for diabetes management.

\cite{pushpanathan2023popular} utilized ChatGPT-3.5 and GPT-4.0, as well as Google's Bard, to evaluate their performance in the ophthalmology department, focusing on addressing queries related to ocular symptoms. The research assessed the clinical applications of these large language models (LLMs) in identifying and managing various ocular conditions. Evaluation metrics included accuracy (graded as poor, borderline, good) and comprehensiveness of responses, alongside the models' self-awareness in terms of their ability to self-check and self-correct. ChatGPT-4.0 emerged as the most accurate, with 89.2\% of its responses rated as 'good', outperforming ChatGPT-3.5 (59.5\%) and Google Bard (40.5\%). All LLMs demonstrated high comprehensiveness scores (4.6 to 4.7 out of 5) but showed subpar to moderate self-awareness capabilities.

\cite{bernstein2023comparison} evaluates the ChatGPT(GPT-3.5) model in the ophthalmology field. Focused on the specialized department of ophthalmology, it addresses the clinical task of providing patient care advice through an online medical forum. Evaluation metrics included the ability to distinguish AI-generated responses from those written by ophthalmologists, presence of incorrect information, alignment with medical consensus, likelihood and extent of harm. The results showed that ChatGPT responses were largely indistinguishable from human-written ones, with similar rates of accuracy regarding correct information, safety, and consensus alignment. The study underlines the potential of LLMs like ChatGPT to generate appropriate ophthalmic advice, comparable to board-certified ophthalmologists.

In evaluating the applications of Large Language Models (LLMs) in the medical industry, \cite{lim2023benchmarking} focused on myopia-related inquiries utilized ChatGPT-3.5, ChatGPT-4.0, and Google's Bard. This research targeted the ophthalmology department, specifically addressing the disease of myopia. The evaluation encompassed various clinical tasks, including pathogenesis, risk factors, clinical presentation, diagnosis, treatment and prevention, and prognosis of myopia. The LLMs' responses were assessed based on accuracy and comprehensiveness by three consultant-level paediatric ophthalmologists using a three-point accuracy scale (poor, borderline, good) and a five-point comprehensiveness scale. ChatGPT-4.0 demonstrated superior performance, with 80.6\% of responses rated as ‘good’ for accuracy, and showed high mean comprehensiveness scores, highlighting its potential in delivering precise and detailed information for myopia care.

In the study\cite{wilhelm2023large} evaluating Claude-instant-v1.0, GPT-3.5-Turbo, Command-xlarge-nightly, and Bloomz. This research specifically targeted the clinical specialties of ophthalmology, orthopedics, and dermatology, encompassing a diverse set of 60 diseases. The evaluation assessed LLMs' ability to generate therapeutic recommendations, focusing on clinical tasks such as the accuracy and safety of treatment advice. The evaluation criteria included the mDISCERN score, correctness, and potential harmfulness of the recommendations. Results showed significant differences among the models in terms of quality and safety across the examined medical fields, with Claude-instant-v1.0 achieving the highest mean mDISCERN score and GPT-3.5-Turbo being noted for the lowest harmfulness rating, indicating the nuanced performance of LLMs in generating clinical recommendations for specific diseases and departments.

\cite{kuroiwa2023potential} assessed ChatGPT(GPT-3.5) focusing on its application within orthopedics, specifically evaluating its ability to self-diagnose common diseases such as carpal tunnel syndrome, cervical myelopathy, lumbar spinal stenosis, knee osteoarthritis, and hip osteoarthritis. The research aimed to measure the model's accuracy and precision in providing diagnoses and recommending medical consultations. Evaluation metrics included correct answer ratios, reproducibility between days and raters using the Fleiss K coefficient, and the extent of medical consultation recommendations. Results indicated variable accuracy with correct answer ratios ranging from 4\% for cervical myelopathy to 100\% for carpal tunnel syndrome. Reproducibility also varied, suggesting inconsistencies in ChatGPT's diagnostic capabilities within the orthopedic domain.

\cite{chervenak2023promise} evaluated the February 2023 version of ChatGPT in the specialized department of reproductive medicine, focusing on fertility-related clinical queries. The clinical tasks assessed included providing information on infertility FAQs from the CDC, completing validated fertility knowledge surveys (Cardiff Fertility Knowledge Scale and the Fertility and Infertility Treatment Knowledge Score), and reproducing the American Society for Reproductive Medicine's committee opinion on "optimizing natural fertility." Evaluations were based on response length, factual content, sentiment analysis, and the accuracy of reproducing key facts. ChatGPT demonstrated comparable performance to established sources in terms of response length, factual content, and sentiment, achieving high percentiles in fertility knowledge surveys and accurately reproducing all key facts from the committee opinion on optimizing natural fertility. This highlights ChatGPT's potential as a relevant and meaningful tool for fertility-related clinical information, though it noted limitations in citing sources and the risk of fabricating information.

\cite{levkovich2023suicide} explores the application of ChatGPT in assessing suicide risk within the mental health department. The experiment utilized ChatGPT versions 3.5 and 4, updated on May 24, 2023, to evaluate their capabilities in a psychiatric context. The clinical task addressed was the assessment of suicide risk, incorporating factors like perceived burdensomeness and thwarted belongingness. Evaluations were conducted based on various indicators such as psychache, suicidal ideation, risk of suicide attempts, and resilience. The results revealed that ChatGPT-4's assessments of suicide risk and suicidal ideation were comparable to those of mental health professionals, highlighting its potential utility in clinical settings. However, it also showed a tendency to overestimate psychache, suggesting areas for further research and improvement.

\cite{liu2023medical} introduces the Medical Multimodal Large Language Model (Med-MLLM), leveraging large-scale pre-training on multimodal medical data. Focused on radiology, specifically chest X-rays (CXR) and Computed Tomography (CT) images\cite{pavlova2022covid, cohen2020covid, cohen2020covid1, liu2021medical, vaya2020bimcv}, Med-MLLM targets diseases like COVID-19 and its variants, including Delta and Omicron. Evaluations span across clinical tasks such as medical reporting, disease diagnosis, and patient prognosis. Performance metrics used include natural language generation metrics (BLEU, ROUGE-L, CIDEr) for reporting, and Accuracy and AUC for diagnosis and prognosis tasks. Remarkably, with just 1\% of labeled data, Med-MLLM demonstrated competitive or superior performance compared to fully-supervised models on various tasks, showcasing its robustness and efficiency in handling rare diseases with limited data.

\cite{goktas2023artificial} explores the utilization of LLMs in clinical tasks such as differential diagnosis and clinical management for diseases specific to these fields. Although not directly assessing performance through metrics, the authors discuss the potential of LLMs to enhance patient engagement, improve diagnostic accuracy, and offer personalized treatment plans. They underscore the necessity for LLMs like ChatGPT 4.0 to address challenges such as data privacy, ethical considerations, and the verification of AI-generated information to ensure their effective and safe use in clinical settings.

\textbf{Supportive Departments Applications: }

In the following section, we delve into the application and evaluation of Large Language Models (LLMs) within supportive departments of the medical field, specifically focusing on radiology and emergency department scenarios. The studies presented explore the utility of models like ChatGPT in radiologic decision-making, diagnostic quizzes in neuroradiology, automatic generation of radiology reports from chest X-rays, and diagnosis and triage in emergency settings. Each research piece offers insights into the capabilities and performance of LLMs across these distinct yet crucial areas of medical practice.

\cite{rao2023evaluating} utilized ChatGPT-3.5 and GPT-4, to explore their capability in clinical decision support within radiology. The research focused on evaluating the models' performance in identifying appropriate imaging services for breast cancer screening and breast pain, crucial areas in radiologic decision-making. The evaluation was conducted against the American College of Radiology (ACR) Appropriateness Criteria, using both open-ended and select-all-that-apply prompt formats. ChatGPT's suggestions were compared with ACR guidelines to assess concordance. Results showed that ChatGPT-3.5 and GPT-4 performed well, with GPT-4 exhibiting a higher average percentage correct in selecting appropriate imaging modalities for breast cancer screening (98.4\%) compared to ChatGPT-3.5 (88.9\%). For breast pain, GPT-4 also outperformed ChatGPT-3.5 in accuracy, demonstrating the potential feasibility of using large language models for radiologic decision-making and indicating a positive trend in model performance with newer versions.

In their study\cite{suthar2023artificial}, the authors evaluated the application of the GPT-4-based ChatGPT (July 20, 2023 version) in the field of neuroradiology, focusing specifically on the radiology department and related diseases of the brain, head and neck, and spine. The research aimed at assessing the model's capability in solving diagnostic quizzes from the "Case of the Month" section of the American Journal of Neuroradiology (AJNR), thereby evaluating its utility in clinical tasks such as differential diagnosis and clinical management. The evaluation criteria included diagnostic accuracy and the validity of differential diagnoses, which were quantified using a five-point Likert scale. The results showcased an overall diagnostic accuracy of 57.86\% across 140 cases, with performance varying by subgroup: 54.65\% for brain, 67.65\% for head and neck, and 55.0\% for spine. This study highlights ChatGPT 4.0's potential as a supportive tool in radiological diagnostics within specific medical domains.

\cite{nicolson2023improving} focuses on the application of LLMs in generating radiology reports from chest X-rays (CXRs), a task critical to diagnostic radiology, a supportive department in healthcare. The research leverages the Convolutional vision Transformer (CvT) ImageNet-21K checkpoint for image encoding and the Distilled Generative Pre-trained Transformer 2 (DistilGPT2) checkpoint for text decoding, aiming to automate report generation in the medical domain of radiology, particularly for diagnosing various conditions depicted in CXRs. The clinical task evaluated is the automatic generation of radiology reports, a pivotal component of clinical management and diagnosis. Evaluation metrics include traditional natural language generation (NLG) metrics (BLEU-4, ROUGE-L, METEOR) and the Clinical Efficacy (CE) metric, focusing on diagnostic accuracy. The model, CvT2DistilGPT2, demonstrates notable improvements over the state-of-the-art, with an 8.3\% increase in CE F-1 score, indicating higher diagnostic accuracy and report quality akin to that of radiologist-generated reports.

\cite{fraser2023comparison} evaluates the performance of ChatGPT versions 3.5 and 4.0 focusing on the Emergency Department (ED) setting. It addresses the clinical tasks of diagnosis and triage for patients with urgent or emergent problems, comparing the LLMs' performance against the symptom checker applications from WebMD and Ada Health, as well as diagnoses and triage recommendations from board-certified ED physicians. Evaluation metrics included diagnostic accuracy—measured by the proportion of matches between the LLMs' diagnoses and final ED diagnoses—and triage accuracy, gauged by the agreement with physician recommendations. ChatGPT 3.5 demonstrated high diagnostic accuracy but a higher unsafe triage rate, whereas ChatGPT 4.0 showed improved triage agreement with physicians but lower diagnostic accuracy. The study underscores the potential and limitations of applying LLMs like ChatGPT in critical clinical settings, highlighting the need for improvements in triage accuracy and thorough clinical evaluation before unsupervised patient use.

\subsubsection{Medical Text Data Processing Application Evaluations}

In a groundbreaking study\cite{peng2023study}, researchers developed GatorTronGPT, a generative large language model (LLM) based on the GPT-3 architecture. This model was trained using a diverse dataset comprising 82 billion words of de-identified clinical text from the University of Florida Health, spanning 126 clinical departments and approximately 2 million patients, as well as 195 billion words of general English text. The evaluation focused on biomedical natural language processing (NLP) tasks, specifically relation extraction and question answering, using six benchmark datasets. GatorTronGPT demonstrated state-of-the-art performance, outperforming existing models with the highest F1 scores in relation extraction for drug-drug interaction, chemical-disease relation, and drug-target interaction, as well as achieving competitive accuracy in biomedical question answering. This highlights GatorTronGPT's potential in processing medical text data, such as clinical notes and electronic health records (EHRs), for advanced NLP applications in the medical industry.

In evaluating the applications of large language models in the medical industry, researchers\cite{chen2023extensive} focused on ChatGPT versions GPT-3.5 and GPT-4. Their comprehensive benchmark study assessed ChatGPT's performance across various medical text data types, including biomedical corpus like article abstracts, clinical trial descriptions, and biomedical questions. The evaluation encompassed a range of data processing tasks such as Named Entity Recognition (NER), Relation Extraction (RE), and Question Answering (QA), among others. Performance metrics included entity-level F1 scores for NER, Micro F1 for RE, and accuracy for QA tasks. Results highlighted that ChatGPT achieved a BLURB score of 58.50, indicating effective yet limited capabilities in biomedical text understanding and generation compared to the state-of-the-art model's score of 84.30.

In a comprehensive evaluation, researchers\cite{jahan2024comprehensive} assessed the performance of four popular LLMs - GPT-3.5, PaLM-2, Claude-2, and LLaMA-2 - across diverse medical text data processing tasks. These tasks, crucial for biomedical text analysis, included Named Entity Recognition (NER), Relation Extraction (RE), Entity Linking, Text Classification, Question Answering, and Text Summarization. Their performance was measured across 26 datasets encompassing various types of medical data such as clinical notes, patient electronic health records (EHRs), and medical research articles. The evaluation focused on metrics like Precision, Recall, F1 scores for NER and RE tasks; Recall@1 for Entity Linking; F1 and Accuracy for Text Classification and Question Answering; alongside ROUGE and BERTScore for Text Summarization. The findings revealed LLMs' strong zero-shot capabilities, particularly in tasks with smaller training datasets, sometimes even outperforming state-of-the-art models that were fine-tuned specifically for those tasks. However, their performance varied across different tasks and datasets, indicating no single LLM consistently outperformed others in all evaluations.

In a retrospective cohort study\cite{mermin2023use} focused on processing patient-initiated electronic health record (EHR) messages, a natural language processing (NLP) framework was developed based on the distilBERT model, a lighter version of BERT optimized for faster computation without significant performance loss. This NLP model was utilized to classify and triage patient communications within the EHR system, specifically targeting messages related to COVID-19. The model's primary tasks involved accurately identifying messages reporting positive COVID-19 test results from a dataset comprising various patient-initiated EHR messages, including test reports and health inquiries. Evaluation metrics centered on classification accuracy, sensitivity, and the macro F1 score. The model demonstrated high effectiveness with a macro F1 score of 94\%, and sensitivities of 85\% for COVID-19 related messages (not reporting a positive test), 96\% for COVID-19 positive test reports, and 100\% for non-COVID-19 communications.

In \cite{alsentzer2023zero}, the researchers evaluated the application of the Flan-T5, a publicly available Large Language Model (LLM), for phenotyping patients with Postpartum Hemorrhage (PPH) from discharge notes within electronic health records (EHR). The dataset consisted of 271,081 discharge summaries. The model's performance was assessed on 24 granular concepts related to PPH, demonstrating its ability to enhance phenotyping accuracy and patient identification for clinical and research purposes. Evaluation metrics included sensitivity, specificity, Positive Predictive Value (PPV), Binary F1 score, and accuracy, with the model achieving strong results, such as a PPV of 0.95 in phenotyping PPH. This signifies the model's high fidelity in identifying relevant medical concepts from unstructured clinical text, showcasing its potential in medical text data processing applications without the need for substantial annotated training data.

In \cite{lituiev2023automatic}, researchers focused on leveraging RoBERTa for processing clinical notes from patients with chronic low back pain (cLBP). The medical data category explored includes clinical notes, encompassing progress notes and emergency department provider notes. The evaluation targeted Named Entity Recognition (NER) and Natural Language Inference (NLI) tasks. Performance metrics such as F1 scores were employed to assess the models, with RoBERTa achieving an F1 score of 84.4\% for NER tasks. Additionally, an entailment model based on RoBERTa demonstrated its capability in identifying social determinants of health (SDoH) through NLI, showcasing notable performance despite the complexity of clinical narratives. This exploration underscores the potential of large language models in processing and extracting valuable information from unstructured medical texts.

In \cite{buonocore2023localizing}, the authors explored the adaptation of BERT and its variant BioBERT, to the Italian language for processing biomedical textual data. They experimented with medical data including machine-translated biomedical abstracts and native Italian medical textbooks, focusing on domain-specific adaptation. The evaluation covered various data processing tasks, such as Named Entity Recognition (NER), Extractive Question Answering (QA), and Relation Extraction (RE), across different biomedical datasets translated into Italian. Performance was measured in terms of Mean Reciprocal Rank (MRR) for language modeling and F1 scores for downstream tasks. The BioBIT model, derived from machine-translated PubMed abstracts, demonstrated significant improvements over the baseline, particularly in NER and RE tasks, with F1 score improvements as high as 5.9\% and 9.4\%, respectively. These results underline the model's effectiveness in biomedical language understanding and processing for the Italian language, addressing the gap in resources for less-resourced languages.

In a study\cite{nowak2023transformer} on leveraging transformer-based models for structuring free-text radiological reports, the authors experiment with BERT, focusing on German chest X-ray reports from intensive care unit (ICU) patients. They evaluate the model's performance on text classification tasks, specifically identifying six radiological findings mentioned in the reports. The evaluation metrics include macro-averaged F1-scores (MAF1) and confidence intervals (CIs). The study compares the efficacy of different pre-training and labeling strategies: an on-site pre-trained model using masked language modeling ($T_{mlm}$) versus a medically pre-trained model ($T_{med}$), both fine-tuned with 'silver' (rule-based) and 'gold' (manually annotated) labels. The highest performance was observed for $T_{mlm}$,gold, achieving a MAF1 of 95.5\% (CI: 94.5–96.3), highlighting the efficiency of custom pre-training combined with manual annotations for medical text data processing applications.

In the research\cite{de2023open} presented, a deep learning system was developed for the task of section identification in clinical narratives, specifically unstructured clinical records such as progress notes written in Spanish. The foundation of this system is a transformer-based pre-trained language model, specifically the BSC's bsc-bio-ehr-es model, which has been fine-tuned on a curated corpus consisting of 1038 annotated clinical documents. The evaluation of the system's performance was conducted using a newly designed B2 metric, focusing on section identification accuracy. This metric allowed for a nuanced assessment of the system's ability to identify and classify sections within the clinical records accurately. The system achieved an average B2 score of 71.3 on the open-source dataset provided for community use, and 67.0 in data scarcity scenarios, showcasing its effectiveness and potential for application in processing clinical narratives within the medical industry.

In a comprehensive analysis\cite{hendrix2023trends} focused on evaluating LLMs utilizing a NLP algorithm for the extraction and categorization of pulmonary nodules from chest CT radiology reports. This investigation specifically processed imaging reports, a vital category of patient electronic health records (EHRs), to evaluate the performance of LLMs in medical text data processing tasks. The NLP algorithm developed for this purpose was assessed for its ability to accurately identify reports containing pulmonary nodules and measure the diameters of these nodules. Evaluation metrics concentrated on sensitivity, specificity, and accuracy, with the algorithm achieving notable results: sensitivity of 94\% for identifying nodule-containing reports, and an accuracy of 93\% in determining the largest reported nodule diameter. This study underscores the efficacy of LLMs in processing and analyzing clinical imaging reports, thereby providing a critical foundation for the early detection and management of lung cancer within the medical field.

In the study \cite{liu2023attention}, a Transformer-based fusion model, ARMOUR, was explored for multimodal data processing in clinical settings. This model was specifically designed to handle structured and unstructured medical data, such as physiological measurements and clinical notes, reflecting its application on diverse medical data types. The evaluation focused on six clinical prediction tasks, including risk prediction and Diagnostic Related Groups (DRG) predictions, to assess the model's performance in handling and processing medical text data alongside structured measurements. Performance evaluations were conducted using metrics like Area Under the Receiver Operating Curve (AUROC) and Area Under the Precision–Recall Curve (AUPRC) for binary classification tasks, and accuracy and Macro F1 for multiclass classification tasks. Results highlighted ARMOUR's superior performance in comparison to unimodal and other multimodal baseline methods, showcasing its effectiveness in integrating structured and unstructured medical data for improved clinical predictions.

In the study \cite{morena2023clinical}, researchers utilized the Savana Manager 3.0 artificial intelligence platform, based on natural language processing (NLP) techniques, for data extraction and analysis. This AI application was deployed on clinical diagnosis data derived from electronic health records (EHRs) provided by the Castilla-La Mancha Regional Healthcare Service (SESCAM) in Spain, covering a period from January 2012 to December 2020. The study focused on processing unstructured clinical information within EHRs, specifically targeting idiopathic pulmonary fibrosis (IPF) diagnosis, to identify patient profiles, diagnostic test utilization, and therapeutic management. Evaluation metrics included precision (P), recall (R), and the F-score, with results indicating high accuracy in identifying IPF cases within the medical records: P=1.0, R=0.862, and F-score=0.926. These findings underscore the effectiveness of applying large language models in medical data processing, particularly in accurately extracting and analyzing patient information from EHRs.

\subsubsection{Medical Research Application Evaluations}
The rapid proliferation of LLMs like ChatGPT in medical research underscores a transformative phase in the intersection of artificial intelligence and healthcare. Although not conducting an empirical evaluation themselves, \cite{temsah2023chatgpt} discussing the use of ChatGPT in the medical field provide essential insights that guide researchers in this domain. These insights, while highlighting the potential enhancements in research methodologies facilitated by LLMs, also underscore the urgent need for empirical studies to validate these technologies in real-world settings. Key points raised in this paper include the urgent call for more empirical research to substantiate the claims of efficiency and effectiveness of LLMs in medical applications. The authors point out that while the citation rate of ChatGPT in PubMed suggests a rapid acceptance within the medical community, a substantial portion of these references are from editorials and commentaries rather than rigorous research studies. This gap indicates a critical need for structured and well-documented empirical research to evaluate the actual impact of LLMs on medical research and practice. Building on the insights highlighted about the potential applications of LLMs in medical research, it is crucial to delve deeper into how these technologies are being empirically evaluated across different facets of scientific inquiry. In this part, we aim to offer overview of evaluations of LLM applied in medical scientific researches, specifically, in three key areas: 1. Biomedical information retrieval and scientific article screening; 2. Modeling and analysis; 3. Writing of scientific article and reference generation. 

\textbf{Retrieval \& Screening: }

MedCPT\cite{jin2023medcpt} stands out by utilizing a Transformer-based model, specifically designed for biomedical information retrieval (IR). MedCPT was experimentally tested on various biomedical IR tasks, demonstrating its capability through zero-shot settings. Performance was assessed on multiple metrics, including document retrieval, article representation, and sentence representation. Notably, MedCPT achieved state-of-the-art results on six different IR tasks, setting new benchmarks in the Benchmarking-IR (BEIR) suite, surpassing traditional sparse retrievers and even larger models such as OpenAI's cpt-text-XL. This highlights its efficacy in generating superior biomedical article and sentence representations, thus underscoring its utility in medical research applications.

\cite{jin2023medcpt} utilized the ChatGPT and GPT-4 APIs to automate the screening of titles and abstracts for clinical reviews. The performance of these large language models (LLMs) was assessed based on accuracy, macro F1-score, sensitivity, and interrater reliability, with k and prevalence-adjusted and bias-adjusted k (PABAK). The results indicated an overall accuracy of 0.91, a macro F1-score of 0.60, a sensitivity for included papers at 0.76, and for excluded papers at 0.91. Additionally, the model demonstrated a high level of agreement with human reviewers (k=0.96 PABAK), highlighting its efficiency and reliability in identifying relevant clinical studies.

\textbf{Modeling \& Analysis: }

In the study\cite{wang2023cmolgpt}, a Generative Pre-trained Transformer-inspired model, named cMolGPT, was utilized to experiment with de novo molecular design targeting specific proteins. This application of a LLMs in medical research was assessed using the generation of SMILES strings for novel compounds. The evaluation metrics included the fraction of valid and unique molecules, fragment similarity, and similarity to the nearest neighbor, among others. Results showed that cMolGPT outperformed baseline models in generating valid and unique molecules, indicating its effectiveness in designing drug-like compounds specific to given targets. These findings highlight the model's potential to contribute significantly to the molecular optimization cycle in drug discovery.

In the study presented in "BioSeq-Diabolo,"\cite{li2023bioseq} researchers utilized NLP-derived semantics analysis methods to evaluate biological sequence similarities, a task analogous to semantics similarity analysis in natural languages. They applied these techniques to assess protein remote homology detection, circRNA-disease associations, and protein function annotation. The evaluation metrics included the Receiver Operating Characteristic Curve (ROC), Precision-Recall Curve (PRC), and Learning to Rank (LTR) integration method. Experimental results demonstrated that the NLP-based methods in BioSeq-Diabolo outperformed other state-of-the-art predictors in these tasks, highlighting the potential of leveraging large language model techniques for complex biological sequence analysis.

RNA-MSM\cite{zhang2024multiple}, an RNA MSA-transformer language model, stands out by leveraging homologous sequence data generated via RNAcmap3. This model is specifically evaluated on tasks related to RNA secondary structure prediction and solvent accessibility prediction. Evaluation metrics include precision-recall curves, F1 scores, and Matthews correlation coefficient (MCC) for secondary structure, along with Pearson correlation coefficient (PCC) and mean absolute error (MAE) for solvent accessibility. The RNA-MSM model demonstrated significant improvements over existing techniques such as SPOT-RNA2, showcasing the robustness and precision of LLMs in deciphering complex biological data and enhancing predictive performance in medical research applications.

In the research\cite{jiang2023health}, the authors developed and evaluated a LLM named NYUTron, based on a BERT-like architecture. NYUTron was specifically tested in the medical domain, focusing on a series of clinical and operational tasks within the NYU Langone Health System. These tasks included predicting 30-day all-cause readmission, in-hospital mortality, comorbidity index, length of hospital stay, and insurance denial predictions. The model's performance was assessed using the Area Under the Curve (AUC) as a primary metric. NYUTron demonstrated AUCs ranging from 78.7\% to 94.9\%, outperforming traditional models by improvements of 5.36\% to 14.7\% across different tasks, showing its effectiveness in leveraging unstructured clinical notes for predictive healthcare analytics.

In the study\cite{lu2023medkpl}, experiments were conducted using the domain-specific Pre-trained Language Model known as CNBERT, which was trained on a large dataset of clinical notes to tailor its capabilities for medical applications. The evaluation was focused on medical text classification tasks. These tasks were assessed using accuracy as the primary performance metric. Results indicated that the CNBERT model outperformed standard NLP methods and even other domain-specific PLMs across various test settings. Specifically, the integration of medical knowledge through the MedKPL framework significantly improved classification accuracy, demonstrating robustness in both multi-classification and binary classification tasks across different medical departments.

In the study\cite{hussain2022artificial}, researchers utilized a novel ensemble-based approach for sentiment analysis on social media data to evaluate public sentiment toward COVID-19 vaccines in the United Kingdom. This ensemble model integrated a hybrid of lexicon rule–based and deep learning–based methods, specifically combining the Valence Aware Dictionary for Sentiment Reasoning (VADER), TextBlob, and the Bidirectional Encoder Representations from Transformers (BERT) model. Evaluation of this application involved monitoring the frequency and nature of adverse effects following immunization (AEFIs) mentioned in social media posts, as well as analyzing sentiment trends related to the vaccines and manufacturers. The study reported an overall positive public sentiment towards vaccines at 58\%, with negative and neutral sentiments at 22\% and 19\% respectively. This ensemble approach demonstrated robust performance in capturing public opinion trends, providing complementary insights for pharmacovigilance and public health policy making. 

\textbf{Writing \& Reference Generation: }

\cite{majovsky2023artificial} utilized the GPT-3 to explore its capability in generating fraudulent scientific articles in the medical field, specifically in neurosurgery. The evaluation focused on the model's ability to mimic the structure and content of authentic medical research papers. Key aspects assessed were linguistic coherence, technical accuracy, and the authenticity of citations and references. The results indicated that while the AI could produce text that superficially resembles a genuine scientific article, including correct formats and a coherent layout, it also generated several incorrect and nonexistent citations, revealing potential flaws in its output when scrutinized by experts.

\cite{doyal2023chatgpt} reviews the application of GPT-3 and GPT-4 in medical writing. It discusses their theoretical use in generating various medical documents, emphasizing the need for oversight by medical professionals to ensure the accuracy and ethical integrity of AI-generated content. Concerns such as bias, misinformation, and privacy are highlighted as significant issues that necessitate ongoing training and regulation of these models. While the discussion is rich in ethical and operational considerations, it does not include results from empirical evaluations but stresses the importance of critical review by experts in actual medical settings.

\cite{tang2023evaluating} investigates the capabilities of GPT-3.5 and ChatGPT in zero-shot medical evidence summarization. The evaluation was conducted on summarization tasks across six clinical domains, utilizing abstracts from Cochrane Reviews. Key aspects assessed were coherence, factual consistency, comprehensiveness, and harmfulness. The results indicated that while LLMs can effectively capture essential information from source documents, they sometimes generate factually inconsistent summaries or potentially harmful content. These findings suggest that while LLMs are promising tools for medical summarization, careful scrutiny is necessary to ensure the reliability and safety of their outputs in medical applications.

\cite{hueber2023quality} discuss the application of the ChatGPT in the medical research field, highlighting its potential in streamlining the creation of scientific documents and bureaucratic texts. However, they caution against the uncritical use of this technology due to observed inaccuracies in content generation, specifically in the realm of scientific citation. The study identifies instances where ChatGPT provided incorrect or fabricated literature references, underscoring the risk of relying on AI-generated data without verification. This analysis suggests that while LLMs like ChatGPT can be useful tools in medical research, their output must be meticulously checked for accuracy and reliability, particularly when used in clinical decision-making or educational settings.

\cite{suppadungsuk2023examining} utilized the GPT-3.5 to assess its effectiveness in identifying references for nephrology literature reviews. The evaluation focused on the accuracy and reliability of reference generation. Key metrics included the existence, completeness, and authenticity of references. Results indicated that 62\% of the references provided by ChatGPT existed, but only 20\% were authentic. The study highlighted a significant number of fabricated (31\%) and incorrect (7\% incomplete) references. Notably, errors were frequent in DOI and link accuracy, suggesting limited reliability of ChatGPT in medical research applications without further verification and cross-checking.

While LLMs show promising enhancements to various aspects of medical research, significant challenges remain that require further investigation. Issues such as data privacy, potential biases, misinformation, and ensuring diversity in data handling need comprehensive assessment. Addressing these concerns is crucial for advancing LLM applications in a manner that is ethically sound and scientifically robust. Future research must continue to evaluate these areas critically, ensuring that advancements in LLM technology align with the overarching goals of improving accuracy, fairness, and security in medical research.

\subsubsection{Medical Education \& Public Awareness Application Evaluations}

\textbf{Medical Education: }

As the medical education landscape seeks innovative ways to enhance learning and knowledge dissemination, LLMs are emerging with significant potential to influence this field. The capabilities of LLMs to understand complex texts, reason through medical scenarios, and generate coherent and contextually relevant responses position them as promising tools in the realm of educational technologies. The application of LLMs in medical education is seen as particularly valuable for a number of reasons: they can act as supplementary resources that provide explanations and case studies, aid in keeping professionals updated with the latest research and evidence-based practices, and offer interactive platforms for learning and assessment. While the tangible impacts of LLMs on medical education are still being explored, their potential to revolutionize learning experiences in this domain is clear. These models can simplify the digestion of complex content through interactive methods, enhancing both the understanding and retention of medical knowledge. Additionally, LLMs offer promising avenues for evaluating academic content and augmenting traditional exam preparation through their advanced reasoning and response generation capabilities. This broad utility signals a shift towards more dynamic and responsive educational environments, where LLMs support a deeper engagement with the material and offer personalized learning experiences. The articles selected for discussion in this section will critically examine these potential applications, assessing the effectiveness and practicality of LLMs in fostering an advanced educational framework for medical students and professionals.

The quality of educational materials is crucial, and while LLMs have the capability to generate high-quality teaching content, it is essential to evaluate their output to prevent the dissemination of inaccuracies or biases to medical professionals. The study\cite{zack2024assessing} utilized GPT-4 to evaluate its application in generating clinical vignettes. These vignettes are essential for training medical students and professionals by simulating patient cases that accurately represent demographic diversity. The evaluation focused on the model's ability to portray the correct demographic distribution of various medical conditions compared to actual US prevalence data. The findings indicated significant biases, as GPT-4 often stereotyped certain demographics in its clinical presentations. For example, it over-represented Black patients in vignettes about sarcoidosis and under-represented other demographics in various conditions. These biases highlight the need for careful implementation and monitoring of LLMs in educational tools to prevent the reinforcement of stereotypes in medical training.

About educational materials, there is a large amount of online medical content, large language models can also be leveraged to evaluate the quality of existing online content, because high-quality educational materials can significantly enhance learning efficiency. The study\cite{golan2023chatgpt} accessed ChatGPT's ability on evaluating online medical content's quality and readability. Specifically, they centered on ChatGPT's proficiency using the DISCERN tool and readability assessments, focusing on shock wave therapy for erectile dysfunction. Evaluations were conducted based on the accuracy of the DISCERN score and various readability indices like the Flesch-Kincaid level and Gunning Fox Index. Results showed a significant discrepancy between ChatGPT's evaluations and those obtained from established human raters and readability tools, indicating that ChatGPT may not yet effectively assess medical content as accurately as human experts and established instruments.

Beyond traditional educational materials, Large Language Models (LLMs) also facilitate the learning process as tools for quick knowledge retrieval and learning support. This study\cite{mago2023potential} utilized the ChatGPT(GPT-3) for educational scenarios within the field of oral and maxillofacial radiology. It assessed the model's capability to provide detailed information on radiographic anatomical landmarks and pathologies. Evaluation was based on a set of 80 questions that covered anatomical landmarks, specific pathologies, and their radiographic features. The model's responses were analyzed and rated using a modified 4-point Likert scale. Results indicated that ChatGPT-3 accurately described radiographic landmarks with a mean score of 3.94 and effectively communicated the characteristic features of pathologies, achieving mean scores of 3.85 for pathologies and 3.96 for their radiographic features. This study underscores the potential of ChatGPT-3 as a supplementary educational tool in medical settings, though it highlighted limitations in detail specificity and the handling of medical abbreviations.

Beyond information retrieval and learning support in educational processes, many researchers have empirically assessed the capabilities of LLMs in answering standardized public examination questions to gauge their understanding and knowledge mastery. These evaluations highlight whether LLMs have the potential to serve as learning assistants. \cite{kaneda2023assessing} utilized versions GPT-3.5 and GPT-4 to answer questions from the 112th Japanese National Nursing Examination (JNNE). The models were assessed based on their ability to provide correct answers across multiple question types, including compulsory, general, scenario-based, and conversation questions. Performance indicators included accuracy rates for each question type. The results showed a significant improvement in accuracy from GPT-3.5 to GPT-4, demonstrating potential benefits of LLMs in medical education settings.

In the study\cite{li2023chatgpt}, ChatGPT(GPT-3) was tested in a mock objective structured clinical examination (OSCE) for the Royal College of Obstetricians and Gynaecologists. The experiment assessed ChatGPT's performance on structured discussion questions related to various obstetrics and gynecology topics. Evaluators judged ChatGPT on its factual accuracy, contextual relevance, communication, information gathering, patient safety, and applied clinical knowledge. The results demonstrated that ChatGPT outperformed average historical human candidate scores, exhibiting rapid and contextually apt responses within the constraints of the examination format.

In the study\cite{brin2023comparing} ChatGPT and GPT-4 were evaluated on their performance in answering USMLE-style questions involving soft skills such as empathy and ethics. Specifically, the models were tasked with 80 multiple-choice questions designed to assess communication, professionalism, and ethical judgment. These questions were sourced from the USMLE website and the AMBOSS question bank. Performance metrics included accuracy and consistency, with GPT-4 demonstrating superior results by correctly answering 90\% of the questions, compared to ChatGPT's 62.5\%, and showing no revisions in its responses, highlighting its robustness and reliability in handling complex medical ethics scenarios.

In addition to standardized test questions, some researchers\cite{kumari2023large} have developed well-designed case vignettes to evaluate LLM capabilities. In a study involving models such as ChatGPT, Google Bard, and Microsoft Bing, 50 hematology-related cases were used to test their application in medical education. These models were assessed on their ability to answer questions across a range of hematology topics. Medical professionals rated the responses on a scale of one to five. Results showed significant performance differences among the models, with ChatGPT scoring the highest, demonstrating its potential effectiveness for educational uses in medicine.

In a comparative study\cite{dhanvijay2023performance}, ChatGPT, Google Bard, and Microsoft Bing were evaluated for their ability to solve case vignettes in physiology education. These models were tasked with responding to 77 validated physiological case vignettes. The evaluation focused on accuracy and appropriateness, using a scale from 0 to 4 structured around the observed learning outcomes. Results showed that ChatGPT outperformed the others, with overall higher accuracy scores, indicating its potential utility in medical education settings.

\textbf{Public Awareness Application: }

In the post-COVID-19 era, with the surge in GPT technology applications, the public is inundated with information of varying quality. Researchers are applying LLMs to sift through this data, identifying authentic information, providing accurate medical insights, and supporting public self-diagnosis efforts. To fully understand the value LLMs bring to these applications, it is crucial to examine the evaluations of these use cases.

Building on the previous discussion of LLMs used to evaluate the quality and readability of online medical content\cite{golan2023chatgpt}, the applicability of LLMs to discern the veracity of information also holds feasibility. The study\cite{alghamdi2023towards} focuses on the detection of COVID-19 related fake news using transformer-based models like BERT and COVID-Twitter-BERT (CT-BERT). The researchers evaluated the effectiveness of these models in a public awareness scenario by discerning the authenticity of information spread on social media platforms. Evaluation metrics included accuracy, precision, recall, and F1 score, with the model CT-BERT coupled with a BiGRU layer achieving an F1 score of 98.5\%, indicating high effectiveness.

If LLMs can accurately discern the veracity of content, it suggests their potential to provide trustworthy and reliable information to the public. Some researchers have explored this capability by evaluating how LLMs deliver medical knowledge, more importantly, assessing their effectiveness in communicating accurate health information. \cite{sezgin2023clinical} utilized LLMs to address frequently asked questions about postpartum depression (PPD), derived from the American College of Obstetricians and Gynecologists (ACOG). Specifically, ChatGPT(GPT-4) and LaMDA (via Bard) were tested alongside Google Search. The evaluation focused on the clinical accuracy of LLM-generated responses to these PPD-related queries. Clinical accuracy was assessed by two board-certified physicians using a GRADE-informed scale, evaluating the appropriateness of responses in reflecting current medical knowledge. Results demonstrated that ChatGPT provided more clinically accurate responses compared to Bard and Google Search, indicating its potential utility in public education by delivering reliable medical information.

\cite{chervenak2023promise} utilized ChatGPT(GPT-3.5) to address fertility-related queries. This application was evaluated in a public awareness context where the AI was tasked to respond to FAQs and clinical queries from reputable sources like the CDC and American Society for Reproductive Medicine. Evaluation metrics included response accuracy, factual correctness, and the ability to replicate expert consensus opinions from medical documents. The results showed that ChatGPT could generate informative and relevant responses, with the model achieving high accuracy in replicating medical advice and facts. However, limitations were noted in source citation and occasional factual inaccuracies.

Building on the potential of LLMs to disseminate medical knowledge to the public, these models also extend their utility to assisting individuals in making preliminary assessments or interpretations of medical conditions based on provided data. In this study\cite{kuroiwa2023potential}, ChatGPT(GPT-3.5) was employed to evaluate their application in public awareness and education within the medical field. The primary task assigned to ChatGPT was to self-diagnose common orthopedic conditions based on symptom descriptions provided by researchers. This application was intended to assess the model’s utility in aiding public self-diagnosis, potentially increasing public awareness about these conditions. The evaluation metrics focused on the accuracy, precision, and the model’s consistency in recommending medical consultations. Results indicated varied correct answer ratios across different conditions, highlighting inconsistent performance. ChatGPT demonstrated high accuracy in diagnosing conditions with localized symptoms such as carpal tunnel syndrome but struggled with conditions having multifocal symptoms like cervical myelopathy. This evaluation showcases the potential and limitations of using LLMs for public education in recognizing symptoms and understanding when to seek professional medical advice.

The European Federation of Clinical Chemistry and Laboratory Medicine (EFLM) Working Group on Artificial Intelligence (WG-AI) employed the ChatGPT to simulate patient scenarios\cite{cadamuro2023potentials}. ChatGPT was tasked with interpreting laboratory test results from fictional clinical cases. These interpretations were evaluated by WG-AI members for their relevance, correctness, helpfulness, and safety. The model's performance in interpreting laboratory results was systematically assessed. While ChatGPT recognized all laboratory tests and could detect deviations from the reference intervals, its interpretations were found to be superficial and not always accurate. The evaluations showed varied results across the quality dimensions, indicating that the model’s utility in providing meaningful medical education or public awareness is limited without further specialized training in medical data.

\subsection{Comprehensive Discussion on Evaluation Methods and Metrics}
This part provides a comprehensive overview of the evaluation methods and metrics used in assessing LLM applications in the medical field, focusing on models, evaluators and comparative experiments, and the diverse range of evaluation indicators.

\subsubsection{Models}\label{subsubsec:Models}
In the ear of AI, many researchers embrace the capable LLMs due their abilities on different tasks, and they could be potentially applied in medical domain. Such that, there are diverse array of LLMs utilized and evaluated in medical applications, ranging from commercial models like GPT-4 and ChatGPT, to open-source frameworks and customized models developed through methods such as fine-tuning on transformer architectures. 

\textbf{GPT series: }

Due to OpenAI's influence and the capabilities of the GPT series, the majority of evaluations of LLM applications in the medical field have focused on GPT models. This includes accessing GPT-3.5 or GPT-4 capabilities through APIs or utilizing these models via the ChatGPT interface, as evidenced by studies such as \cite{liu2023utility, majovsky2023artificial, grewal2023radiology, rao2023assessing, chen2023extensive, sun2023ai, guo2024automated, bernstein2023comparison, suppadungsuk2023examining, kuroiwa2023potential, chervenak2023promise, zack2024assessing, cadamuro2023potentials, li2023chatgpt, hueber2023quality, golan2023chatgpt, suthar2023artificial, mago2023potential, kaneda2023assessing, rao2023evaluating, levkovich2023suicide, fraser2023comparison, brin2023comparing, goodman2023accuracy, tang2023evaluating}.

\textbf{Other Commercial Models: }

Besides, commercial models such as Claude\cite{anthropic2024introducing}, Bard\cite{manyika2023overview}, PALM\cite{anil2023palm}, LaMDA\cite{thoppilan2022lamda}, and Command\cite{cohere2023command} are also prevalent in medical evaluations. Some studies extend their assessments beyond the GPT series to include these models, facilitating a more comprehensive evaluation. By comparing the performance of multiple commercial models, these studies provide a broader understanding of the differences between each model\cite{sezgin2023clinical, pushpanathan2023popular, lim2023benchmarking, jahan2024comprehensive, wilhelm2023large, kumari2023large, dhanvijay2023performance}.

\textbf{Open-Source Models: }

In addition to commercial LLMs, numerous open-source models in general domain like BERT\cite{devlin2018bert}, GPT2\cite{radford2019language}, RoBERTa\cite{liu2019roberta}, LLaMA\cite{touvron2023llama}, Llama 2\cite{touvron2023llama2}, ALBERT\cite{lan2019albert}, T5\cite{raffel2020exploring}, FLAN-T5\cite{chung2024scaling}, BLOOMZ\cite{muennighoff2022crosslingual}, PRIMERA\cite{xiao2021primera} and DistilBERT\cite{sanh2019distilbert}, and in specific domain like BlueBERT\cite{peng2019transfer}, BioGPT\cite{luo2022biogpt}, BioBART\cite{yuan2022biobart}, BioBERT\cite{lee2020biobert}, ClinicalBert\cite{alsentzer2019publicly}, ProtTrans\cite{elnaggar2007prottrans}, BioSeq-BLM\cite{li2021bioseq}, PubMedBERT\cite{gu2021domain}, SciBERT\cite{beltagy2019scibert}, bsc-bio-ehr-es\cite{carrino2022pretrained} are also utilized in application and evaluation researches\cite{jahan2024comprehensive, wilhelm2023large, mermin2023use, li2023bioseq, nicolson2023improving, alsentzer2023zero, lituiev2023automatic, buonocore2023localizing, nowak2023transformer, de2023open, hussain2022artificial}. While commercial models are generally designed for general domains and have improved in specialized capabilities over iterations, they still exhibit gaps in domain-specific performance. Open-source models remain vital, offering significant value to the medical field. These models allow developers to customize the architecture and data specifically for their application scenarios, enhancing suitability and opening up new possibilities for LLM applications. Studies employing open-source LLMs demonstrate their practical utility and flexibility in addressing specific medical tasks\cite{singhal2023large, peng2023study, jin2023medcpt, jiang2023health, liu2023medical, lu2023medkpl, zhang2024multiple}. Some more innovative researches have specifically developed LLMs tailored for medical applications, addressing specific requirements and scenarios within the medical industry. Plus, these researches also conduct horizontal comparisons among multiple LLMs.

\subsubsection{Evaluators and Comparative Experiments}

We explore the diversity in evaluators ranging from expert assessments to automated metrics, and the various comparative experimental setups utilized in the part. It will discuss about how the different evaluators contribute to the evaluation of LLMs and the comparative experiments employed.

\textbf{Evaluators: }

In the evaluation of LLMs, there are roughly three types of evaluators: human experts, automated calculations based on predefined metrics, and AI-driven automated assessments. Most studies currently employ the first two methods. 

\begin{itemize}

\item \textbf{Human Evaluator:} Human experts often serve as the primary evaluators \cite{liu2023utility, majovsky2023artificial, peng2023study, grewal2023radiology, rao2023assessing, sun2023ai, sezgin2023clinical, pushpanathan2023popular, bernstein2023comparison, suppadungsuk2023examining, wilhelm2023large, rao2023evaluating, kuroiwa2023potential, chervenak2023promise, zack2024assessing, cadamuro2023potentials, kumari2023large, li2023chatgpt, singhal2023large, lim2023benchmarking, hueber2023quality, goodman2023accuracy, suthar2023artificial, mago2023potential, dhanvijay2023performance}. 
For example, in\cite{liu2023utility}, various domain-specific experts such as general internal medicine physicians, radiologists, and transplant hepatologists evaluated the tasks performed by ChatGPT. These evaluations were based on standards such as diagnostic accuracy, clinical appropriateness criteria from the American College of Radiology (ACR), and reliability metrics like Cronbach’s alpha. These experts assessed the model's performance in generating differential diagnoses, medical documentation, and answers to medical queries. 
Two independent experts in radiology evaluated the performance of ChatGPT-3.5 and ChatGPT-4 in \cite{rao2023evaluating}. The evaluation was also based on the ACR appropriateness Criteria, which serve as the standard for imaging recommendations in breast cancer screening and breast pain management. These criteria assess the clinical utility of diagnostic modalities given specific patient presentations. 
An evaluation of ChatGPT's performance on clinical vignettes was conducted by independent scorers\cite{rao2023assessing}. These scorers assessed the LLM's accuracy in generating differential diagnoses, diagnostic tests, final diagnoses, and management strategies. The reference standard for this evaluation was the Merck Sharpe \& Dohme (MSD) Manual, which is widely accepted in the medical field. The accuracy was determined by comparing ChatGPT’s responses against these established medical guidelines.
These experts assess LLM performance on various tasks based standards or their professional experience. This widely use underscores the ongoing critical role of human expertise in integrating LLMs into medical applications, a trend that is likely to continue for the foreseeable future. Although human experts offer invaluable insights, their evaluations are challenging to scale up due to the scarcity of expert resources and the inherent inefficiencies of manual assessment. Additionally, the potential for subjective biases in expert evaluations cannot be ignored. As a result, when considering human experts as evaluators, it may be beneficial to incorporate consistency measures across multiple experts' assessments, as discussed in \cite{guo2024automated}, to ensure a more robust and objective evaluation.

\item \textbf{Automated Metrics Calculation:} Another widely used method of evaluation employs automated metrics calculations, which typically rely on predefined gold standards provided by human experts. These metrics are commonly used in traditional natural language processing tasks such as NER and Relation Extraction, or are derived from real clinical case data, or specific medical examinations. These studies have employed this method to evaluate LLMs' performance. \cite{peng2023study, wang2023cmolgpt, alghamdi2023towards, chen2023extensive, sun2023ai, guo2024automated, kaneda2023assessing, jahan2024comprehensive, mermin2023use, li2023bioseq, nicolson2023improving, zhang2024multiple, jin2023medcpt, jiang2023health, levkovich2023suicide, fraser2023comparison, brin2023comparing, liu2023medical, alsentzer2023zero, golan2023chatgpt, lituiev2023automatic, buonocore2023localizing, nowak2023transformer, de2023open, hendrix2023trends, lu2023medkpl, liu2023attention, tang2023evaluating}
While this approach is widely adopted for its efficiency, it often lacks depth in assessing the quality of content, such as coherence and logical consistency of the descriptions, particularly as more content is generated by LLMs in contemporary applications. Similarly, consistency should be taken into consideration as well. \cite{lituiev2023automatic} used Cohen's Kappa to evaluate the inter-rater agreement between different annotators.   

\item \textbf{AI Evaluator:} A minority of studies have utilized AI models as automated evaluators. This method is not yet widely used but is gaining interest as LLMs increasingly excel in understanding text content, reasoning, and aligning with human preferences, which could help them perform deeper evaluations that the automated metrics evaluation methods mentioned above may miss. 
In \cite{majovsky2023artificial}, two AI detection tools were employed: AI Detector by Content at Scale and AI Text Classifier by OpenAI. The AI Detector predicts AI generation likelihood by analyzing word choice patterns. The AI Text Classifier categorizes content on a scale from "very unlikely" to "likely" AI-generated. Both tools are used to assess whether the text produced by ChatGPT appears to be human or machine-generated.
An AI's automatic evaluation of other LLMs' responses was performed using GPT-4\cite{wilhelm2023large}. This evaluation utilized a modified version of the DISCERN instrument, termed mDISCERN, to assess the quality of AI-generated therapeutic recommendations. The mDISCERN criteria focus on the correctness and potential harmfulness of the content, aligning with established health information standards.
In \cite{chervenak2023promise}, the evaluation of ChatGPT was performed by using the Python library TextBlob for sentiment analysis. TextBlob assesses the polarity (ranging from -1 to +1, where negative values indicate negative sentiment, and positive values indicate positive sentiment) and subjectivity (ranging from 0 to 1, where values closer to 0 indicate objectivity) of the text. The assessment involved comparing factual content and sentiment of ChatGPT's responses against authoritative sources like the CDC and established medical guidelines.
However, LLMs are not authoritative experts, so starting with a supportive role in evaluation might be more feasible. To ensure the credibility of LLMs' evaluations, considering independent assessments from multiple LLM providers may be necessary to achieve a consensus, and the consistency of evaluations across different models must be considered. Additionally, it's crucial that the evaluations by a single model across various datasets remain fair, uniform, and unbiased, which could be referred to \cite{zack2024assessing}.

\end{itemize}

\textbf{Comparative Experiments: }

In addition to evaluator, this part outlines how research validates the capabilities of models through different comparative setups, including comparisons between different LLMs, traditional NLP algorithms, rule-based systems, and existing platforms. Additionally, some studies assess models in isolation against benchmarks or through expert evaluation, while others compare model performance directly with human capabilities on identical tasks.

\begin{itemize}

\item \textbf{Compare to LLMs: }As the prevalence of LLMs grows, many researchers focus on comparative studies involving these models to assess performance. This involves comparisons among LLMs to evaluate their capabilities across various tasks, as documented in studies \cite{liu2023utility, peng2023study, alghamdi2023towards, chen2023extensive, sun2023ai, sezgin2023clinical, kaneda2023assessing, pushpanathan2023popular, jahan2024comprehensive, mermin2023use, wilhelm2023large, li2023bioseq, rao2023evaluating, nicolson2023improving, zhang2024multiple, kumari2023large, singhal2023large, jiang2023health, levkovich2023suicide, brin2023comparing, liu2023medical, lim2023benchmarking, goodman2023accuracy, buonocore2023localizing, nowak2023transformer, dhanvijay2023performance, de2023open, lu2023medkpl, tang2023evaluating}. Notably, representative works such as \cite{peng2023study, alghamdi2023towards, chen2023extensive, jahan2024comprehensive, nicolson2023improving, singhal2023large} have conducted extensive comparisons across multiple LLMs on a variety of tasks. (For specific model classifications and descriptions, refer to \ref{subsubsec:Models}, while here we focus solely on the comparative experimental aspect.) 
\cite{peng2023study} involved a comparison of the GatorTronGPT model with several large language models including GPT-2, REBEL, REBEL-pt, BioGPT, PubMedBERT, BioELECTRa, BioLinkBERT, and Galactica. The comparisons were conducted across tasks in relation extraction and question answering. 
The comparative experiments in \cite{alghamdi2023towards} involved BERT, RoBERTa, DistilBERT, and COVID-Twitter-BERT (CT-BERT), which were further fine-tuned and evaluated on COVID-19 fake news detection tasks.
In \cite{chen2023extensive}, experiments involved the ChatGPT(GPT-3.5) and baseline models including PubMedBERT, BioLinkBERT-Base, and BioLinkBERT-Large, all evaluated on the BLURB benchmark across various biomedical NLP tasks.
In \cite{jahan2024comprehensive} GPT-3.5, PaLM-2, Claude-2, and LLaMA-2 are involved. They were compared against state-of-the-art fine-tuned models such as BioGPT, BioBART, and BioBERT across various biomedical tasks.
\cite{singhal2023large} compare their models, Flan-PaLM and Med-PaLM, against severalLLMs including Galactica, PubMedGPT, BioGPT, and DRAGON. 

\item \textbf{Compare to Specific Algorithms and Systems: }In addition to LLMs, the subjects of comparison may also include publicly available platforms or systems, earlier natural language processing algorithms, machine learning algorithms, and rule-based systems. 
\textbf{For the scenario of clinical application}, \cite{fraser2023comparison} compared the diagnostic and triage performance of the LLMs like ChatGPT 3.5 and ChatGPT 4.0 with the widely used WebMD symptom checker and the Ada Health symptom checker. 
\cite{alsentzer2023zero} compared the performance of the Flan-T5 language model to regular expressions specifically constructed for extracting postpartum hemorrhage (PPH)-related concepts from clinical discharge summaries.
\cite{lu2023medkpl} compared the MedKPL framework against traditional NLP models such as LSTM, LSTM with attention, and CNN, as well as pre-trained language models including BERT and DKPLM, employing various training approaches like fine-tuning and prompt learning.
\textbf{In the field of bioinformatics}, studies \cite{wang2023cmolgpt, zhang2024multiple} compared commonly used algorithms and models specific to their respective domains. Specifically, in \cite{wang2023cmolgpt}, the cMolGPT model was compared against several baseline models including Hidden Markov Model (HMM), N-gram generative model, combinatorial generator, character-level recurrent neural network (CharRNN), SMILES variational autoencoder (VAE), adversarial autoencoder (AAE), junction tree VAE (JTN-VAE), and latent vector-based generative adversarial network (LatentGAN). And \cite{zhang2024multiple} compares their RNA-MSM model with existing methods including the RNA-FM, a BERT-based RNA language model, traditional folding-based techniques like RNAfold and LinearPartition, and other RNA structure-trained secondary structure predictors such as SPOT-RNA and SPOT-RNA2. Additionally, solvent accessibility prediction comparisons involve RNAsnap2 and M2pred.
\textbf{In other specialized areas}, there is no shortage of studies comparing LLMs with traditional algorithms or machine learning methods. 
\cite{alghamdi2023towards} not only use language models but also classic machine learning models like logistic regression, support vector machines, naive Bayes, random forests, and XGBoost. 
The comparison experiments in \cite{jin2023medcpt} involved comparing the MedCPT model against various models and algorithms, including sparse retrievers like BM25 and docT5query, dense retrievers such as DPR, ANCE, TAS-B, and Contriever, and language model retrievers like Google's GTR series and OpenAI's cpt-text series. 
\cite{lituiev2023automatic} compares several models and systems including a rule-based NER system (cTAKES with custom configurations), a convolutional neural network (CNN) implemented in spaCy, a RoBERTa-based NER model, a hybrid model combining rule-based extraction with logistic regression, and a pre-trained RoBERTa entailment model.

\item \textbf{Isolation: }Unlike the comparative studies mentioned earlier, some research simply assesses models in isolation, which means the evaluation is not in the form of comparison among models but merely on one model. Usually, they are conducted by utilizing annotated datasets or benchmarks to derive performance metrics, or through evaluations conducted by human experts following specific standards. In these studies, most of them are evaluated by human evaluator\cite{liu2023utility, majovsky2023artificial, grewal2023radiology, rao2023assessing, sun2023ai, suppadungsuk2023examining, kuroiwa2023potential, zack2024assessing, cadamuro2023potentials, hueber2023quality, suthar2023artificial, mago2023potential}. Specifically, 
The evaluation in \cite{grewal2023radiology} was conducted by an experienced radiologist, co-author HG, who assessed the performance of the GPT-4 model in tasks such as generating radiology reports and templates.
Clinical nutrition experts assessing the performance of the ChatGPT model on answering common questions related to ketogenic diet therapy in \cite{sun2023ai}. 
In \cite{suppadungsuk2023examining}, evaluation of the authenticity and accuracy of references provided by ChatGPT in the field of nephrology was conducted by the research team using databases such as PubMed, Google Scholar, and Web of Science.
Seven members of the EFLM Working Group on Artificial Intelligence assessed the performance of the ChatGPT model in interpreting laboratory test results across various clinical cases \cite{cadamuro2023potentials}.
\cite{hueber2023quality} manully evaluates the performance of ChatGPT, focusing on its ability to generate accurate scientific citations and maintain content quality in tasks related to medical documentation and editorial writing in rheumatology.
Three fellowship-trained neuroradiologists who assessed the performance of the GPT-4-based ChatGPT model on the task of solving diagnostic quizzes from the "Case of the Month" feature in the American Journal of Neuroradiology (AJNR) \cite{suthar2023artificial}.
An experienced oral and maxillofacial radiologist assessed the performance of the ChatGPT-3 for its ability to answer questions related to anatomical landmarks, oral and maxillofacial pathologies, and their radiographic features in oral and maxillofacial radiology \cite{mago2023potential}.
Some of studies that are without comparison are evaluated with automated metrics calculation. In \cite{guo2024automated}, datasets used in the study were organized by the authors, and the experiment is accessing LLMs' ability on include or exclude scientific papers for systematic review. Evaluation dataset was annotated by a medical student under supervision and an experienced radiologist, and further independently annotated by another experienced radiologist for the test set in \cite{hendrix2023trends}. The performance of a natural language processing algorithm developed to detect pulmonary nodules described in radiology reports was assessed on this dataset. 

\item \textbf{Compare to Human Experts: }Before LLMs are widely applied in the medical domain, it is essential to validate their capabilities against those of professionally trained doctors and experts to understand the differences in performance. Some researches' comparison are conducted between model and human experts. 
\cite{liu2023utility} describes several comparisons between ChatGPT and human experts: Hirosawa et al. found ChatGPT-3 slightly less accurate in generating differential diagnoses than general physicians (93.3\% vs. 98.3\%). Rao et al. reported that ChatGPT achieved a 71.7\% accuracy rate in clinical vignettes compared to the Merck Manual. Liu et al. observed mixed ratings on recommendations by ChatGPT compared to expert-generated ones, with differences in understandability and relevance.
In \cite{bernstein2023comparison}, responses generated by the ChatGPT chatbot were compared with those from AAO-affiliated ophthalmologists on an online forum. An expert panel of 8 board-certified ophthalmologists evaluated 200 paired responses, blinded to whether answers were AI-generated or human-written. They assessed on accuracy, appropriateness, and safety. The panel correctly distinguished between AI and human responses 61.3\% of the time, indicating that the responses from models and human experts are distinguishable. Besides, it describes the other several comparisons between ChatGPT and human experts: Hirosawa et al. found ChatGPT-3 slightly less accurate in generating differential diagnoses than general physicians (93.3\% vs. 98.3\%). Rao et al. reported that ChatGPT achieved a 71.7\% accuracy rate in clinical vignettes compared to the Merck Manual. Liu et al. observed mixed ratings on recommendations by ChatGPT compared to expert-generated ones, with differences in understandability and relevance.

\end{itemize}

\subsubsection{Evaluation Metrics}

\textbf{Correctness: }

In medical field, correctness is the most important aspect that the researchers focus on. Such that, correctness metrics, such as accuracy, precision etc are most commonly used in evaluations. 

\begin{itemize}

\item \textbf{For clinical application, }accuracy is a widely used metric usually for tasks of diagnosis, prognosis, decision making, risk prediction etc. 
In \cite{rao2023assessing}, LLMs' clinical decision support capabilities were evaluated using accuracy, specifically overall accuracy and accuracy by question type: differential diagnosis , diagnostic testing, final diagnosis , and management. Accuracy was calculated by the proportion of correct responses compared to standardized clinical vignettes from the Merck Sharpe \& Dohme Clinical Manual. 
\cite{suthar2023artificial} evaluated the performance of LLM in diagnosing cases from the AJNR's "Case of the Month" using various correctness metrics. The overall diagnostic accuracy, subgroup accuracies for brain cases, neck cases and spine cases. Additionally, a Likert scale (1 to 5) was used to assess diagnostic probability, with scores of 4 and 5 indicating satisfactory performance. 
\cite{fraser2023comparison} evaluated the performance of ChatGPT, WebMD, and Ada Health SCs in diagnosing and triaging urgent or emergent clinical cases. Diagnosis accuracy was assessed by the proportion of system-generated diagnoses matching the final ED diagnosis, while triage accuracy was determined by the agreement of recommendations with independent physician reviews. 
In \cite{zack2024assessing}, the authors evaluated GPT-4 using accuracy metrics to assess its performance in four clinical applications: medical education, diagnostic reasoning, clinical plan generation, and subjective patient assessment. They compared GPT-4’s estimates of demographic distribution of medical conditions against true US prevalence estimates and analyzed differential diagnosis and treatment planning across demographic groups. Statistical tests such as the $\chi^2$ test of independence, Mann-Whitney tests, and logistic regression were employed to determine significant differences and biases in GPT-4’s responses.
In \cite{lu2023medkpl}, they used accuracy to calculate the proportion of correctly classified instances out of the total instances. It is applied to evaluate the performance of the Medical Knowledge-enhanced Prompt Learning (MedKPL) framework and its domain-specific pre-trained language model (CNBERT) on clinical note classification tasks. The experiments are conducted on two medical EHR datasets, focusing on multi-classification and binary classification tasks to enhance the transferability and robustness of the models in clinical diagnosis.
\cite{liu2023attention} evaluates model for clinical prediction tasks using structured measurements and unstructured text. They uses accuracy for multiclass classifications. It assesses model correctness by measuring the ability to distinguish between different classes accurately, specifically for Diagnostic Related Group (DRG) predictions. 

In addition to accuracy, there are other metrics used for clinical application. \cite{jiang2023health} employed other correctness evaluation metrics such as true positive rate (TPR), false positive rate (FPR), and precision to assess the performance of the NYUTron model on clinical applications. TPR measures the proportion of actual positives correctly identified, FPR the proportion of negatives incorrectly identified as positives, and precision the proportion of true positives among the predicted positives. They were applied to evaluate the model’s effectiveness in tasks like 30-day readmission prediction, in-hospital mortality prediction, comorbidity index imputation, length of stay prediction, and insurance denial prediction.
Similarly, \cite{mermin2023use} used precision to evaluate. The model was applied to classify patient-initiated EHR messages for timely COVID-19 case identification and antiviral treatment facilitation.

\cite{liu2023utility} and \cite{rao2023evaluating} employed other correctness metrics to evaluate LLMs in clinical applications. 
Metrics used in \cite{liu2023utility} included correct diagnosis rate (percentage of correct diagnoses), overall accuracy rate (percentage of accurate responses), SATA (select all that apply, average correct rate), OE score (open-ended, score out of 2), and accuracy of clinical letters (median accuracy and humanness scores). These metrics were applied to assess differential diagnosis lists, clinical decision-making, cancer screening, and the generation of patient clinical letters. 
\cite{rao2023evaluating} used OE scores and SATA in evaluation as well, specifically for clinical decision support in radiology. OE scores were based on the appropriateness of a single imaging procedure, while SATA scores measured the proportion of correct imaging modality selections. The models were assessed on identifying appropriate imaging services for breast cancer screening and breast pain, compared to the ACR Appropriateness Criteria. 

\item \textbf{For specific tasks, }such as NLP, NLI (natural language inference) etc, precision is more frequently used. 
In \cite{peng2023study}, the researchers evaluated GatorTronGPT using precision for biomedical relation extraction tasks (e.g., drug-drug interaction, chemical-disease relation, drug-target interaction). 
\cite{jahan2024comprehensive} evaluates LLMs using precision, accuracy, and Recall@1 metrics. Recall@1 is not quite often used. They employed it to assess the accuracy of top-ranked predictions. 
\cite{alsentzer2023zero} evaluated model for phenotyping postpartum hemorrhage (PPH) using discharge notes. Precision and positive predictive value (PPV) were the main correctness metrics. PPV is the proportion of true positive cases among the identified cases. 
The correctness of models applied to extract social determinants of health (SDoH) from clinical notes of chronic low back pain (cLBP) patients was evaluated using precision as a key metric in \cite{lituiev2023automatic}. The evaluation encompassed various named entity recognition (NER) systems, including a rule-based system (cTAKES), RoBERTa NER, and a hybrid model.
\cite{morena2023clinical} utilized the Savana Manager 3.0 AI platform to evaluate the clinical profile and therapeutic management of patients with idiopathic pulmonary fibrosis. Correctness was assessed using precision for identifying IPF cases within electronic health records. 
\cite{hendrix2023trends} evaluated the correctness of a NLP algorithm used for detecting pulmonary nodules in radiology reports. They used accuracy. The algorithm was applied in the clinical context of identifying and managing pulmonary nodules in chest CT scans for early-stage lung cancer detection.
Mean reciprocal rank (MRR) and pseudo-perplexity (PPPL) are used as evaluation metrics in \cite{buonocore2023localizing}. MRR measures the ranking quality by evaluating the top five predicted tokens for manually curated medical sentences, while PPPL evaluates model performance in masked language modeling. These metrics assess the correctness of the pretrained models for biomedical domain adaptation.

\item \textbf{For medical examinations, }the most often used correctness metric is accuracy. 
In \cite{sun2023ai}, the accuracy rate was used to assess the performance of ChatGPT and GPT 4.0. It was calculated as the percentage of correctly answered questions in the Chinese Registered Dietitian Examination. 
\cite{kaneda2023assessing} assessed the correctness of GPT-3.5 and GPT-4 using accuracy rates and total accuracy rates based on responses to the 112th Japanese National Nursing Examination (JNNE). Correct answer rates were calculated for compulsory, general, and scenario-based questions, comparing performance between the two versions. 
Similarly, \cite{brin2023comparing} applied accuracy in evaluation. It is the percentage of correctly answered questions out of the total 80 USMLE-style soft skills questions, sourced from the USMLE website and AMBOSS question bank. 

\item \textbf{For question answering, }accuracy is often used as well. Some researches evaluate correctness on public QA benchmarks and datasets. 
\cite{chen2023extensive} used accuracy-related metrics to evaluate ChatGPT's performance in biomedical NLP tasks, including question answering (QA) with accuracy as a primary measure. The accuracy metric was applied to the PubMedQA and BioASQ datasets to determine the correctness of the model's responses in clinical applications.
In \cite{singhal2023large}, LLMs, including PaLM, Flan-PaLM, and Med-PaLM, were evaluated using correctness metric accuracy. It was measured on multiple-choice datasets like MedQA, MedMCQA, and PubMedQA. 

Moreover, some other researches conducted evaluations with respect to well-designed standards.
\cite{pushpanathan2023popular} assessed the accuracy of ChatGPT-3.5, ChatGPT-4.0, and Google Bard in responding to 37 common ocular symptom inquiries. Accuracy was evaluated using a three-tier grading system ('poor,' 'borderline,' 'good') based on the responses' potential to mislead or cause harm. Responses were independently graded by three ophthalmologists, and the final accuracy rating was determined by majority consensus. 
\cite{kumari2023large} utilized accuracy scores ranging from 1 to 5 to evaluate the correctness of LLM responses. Scores were given by three raters based on the precision of the answers: 5 (highly accurate) to 1 (inaccurate). The average scores were compared using Friedman’s test with Dunn’s post-hoc analysis. This evaluation focused on the application of LLMs in solving hematology-related question-answering tasks. 
\cite{wilhelm2023large} evaluated the correctness of LLMs using the mDISCERN score to assess the quality of medical information generated by LLMs. Correctness was determined by measuring the presence of false information in the responses. The evaluation involved therapeutic recommendations for 60 diseases across ophthalmology, dermatology, and orthopedics. Physicians rated the accuracy based on practical clinical knowledge, UpToDate, and PubMed, and conducted ANOVA and pairwise t-tests to analyze differences in content quality among the models.
\cite{kuroiwa2023potential} used "correct answer ratio" and "error answer ratio" to evaluate the accuracy of ChatGPT in diagnosing five common orthopedic conditions. Correct answer ratio is the percentage of accurate diagnoses among total responses, and error answer ratio is the percentage of incorrect diagnoses. The conditions evaluated were carpal tunnel syndrome, cervical myelopathy, lumbar spinal stenosis, knee osteoarthritis, and hip osteoarthritis. These metrics assessed ChatGPT's performance in providing accurate self-diagnoses based on standardized symptom-based questions over a 5-day period .
\cite{lim2023benchmarking} evaluated the accuracy of ChatGPT-3.5, ChatGPT-4.0, and Google Bard in answering 31 common myopia-related questions. Accuracy was assessed by three paediatric ophthalmologists using a three-point scale (poor, borderline, good). The final rating for each response was determined by majority consensus. 
In \cite{chervenak2023promise}, correctness was evaluated by comparing the number of factual statements in responses to infertility FAQs from the CDC. The evaluation metrics included the total number of factual statements, the rate of incorrect statements, and whether references were cited. ChatGPT's answers were analyzed for factual accuracy, and any incorrect statements were noted. 
\cite{bernstein2023comparison} evaluated the correctness of ChatGPT-generated ophthalmology advice using the metric of presence of incorrect or inappropriate material. The evaluation was conducted by a panel of 8 board-certified ophthalmologists, who independently assessed whether the answers contained incorrect or inappropriate information. The comparison was between 200 AI-generated responses and human-written responses from the Eye Care Forum. 
\cite{cadamuro2023potentials} assessed ChatGPT's correctness in interpreting laboratory test results using an ordinal scale from 1 (very low) to 6 (very high). Correctness was defined as the scientific and technical accuracy of ChatGPT's interpretations based on best available medical evidence and laboratory medicine practices. Evaluators rated the AI's responses for individual and overall test interpretations. 
In \cite{goodman2023accuracy}, the correctness of responses was evaluated using a 6-point Likert scale for accuracy, where 1 indicated "completely incorrect" and 6 indicated "completely correct." Physicians across 17 specialties generated 284 medical questions, which were rated for accuracy based on medical expertise. The study focused on assessing the chatbot's performance in answering physician-developed medical queries.

\item \textbf{For information retrieval and reference support, }correctness metrics like accuracy, precision, MAP, NDCG etc are used. 
In \cite{jin2023medcpt}, MedCPT is evaluated using Mean Average Precision (MAP) and Normalized Discounted Cumulative Gain (NDCG) metrics. MAP measures the mean of the precision scores after each relevant document is retrieved, reflecting the precision of the top results. NDCG assesses the ranking quality by considering the positions of relevant documents, emphasizing higher-ranked documents. 
\cite{guo2024automated} evaluated the correctness of GPT models for screening clinical review titles and abstracts using accuracy. It was calculated as the proportion of papers correctly identified by the model compared to human reviewers. The evaluation was based on ground truth labeling by two independent human reviewers and was applied to over 24,000 titles and abstracts from six different medical review data sets.
\cite{suppadungsuk2023examining} assessed the correctness of ChatGPT in identifying references for literature reviews in nephrology. It evaluated six components: authors, reference titles, journal names, publication years, digital object identifiers (DOIs), and reference links. References were categorized as authentic if all six components were correct, fabricated if all were false, incomplete if any component was missing, and partially correct if some but not all components were accurate. The validation was conducted using PubMed, Google Scholar, and Web of Science to ensure reliability.
In \cite{hueber2023quality}, the evaluation metrics focus on the accuracy and quality of citation data generated by AI. The correctness of citation data is assessed by verifying the existence and authenticity of referenced literature. AI-generated citations are cross-checked against independent literature databases such as PubMed, Scopus, and Web of Science. 
In addition to scientific paper retrieval and reference support, evaluation of LLMs applied in fake new recognizing can also be conducted using accuracy and precision as correctness metrics. 
In \cite{alghamdi2023towards}, accuracy and precision were used as evaluation metrics to assess the correctness of various models for COVID-19 fake news detection. Accuracy measures the overall correctness of the model's predictions, while precision calculates the proportion of true positive results among all positive results predicted. These metrics evaluated models like BERT, CT-BERT, and their variations with CNN and BiGRU layers.

\end{itemize}

\textbf{Completeness: }

On the other hand, the completeness metric is an essential indicator for evaluating the performance of LLMs. Complementing the correctness metric, it reflects the comprehensiveness of the content provided by LLMs, ensuring that the generated content covers all necessary aspects of specific inputs. 
Many studies that focus on the correctness metric mentioned above also employ the completeness metric to ensure thorough evaluation. This includes works on information organization\cite{alghamdi2023towards, guo2024automated, suppadungsuk2023examining}, various NLP tasks\cite{jahan2024comprehensive, alsentzer2023zero, lituiev2023automatic, hendrix2023trends, morena2023clinical}, QA applications\cite{pushpanathan2023popular, singhal2023large, lim2023benchmarking, goodman2023accuracy, zack2024assessing}, and clinical applications\cite{mermin2023use, jiang2023health}.

Recall (sensitivity) was used to evaluate COVID-19 fake news detection models in \cite{alghamdi2023towards}. Recall calculates the proportion of true positive results correctly identified by the model, crucial for assessing the completeness of information detection. 
\cite{guo2024automated} used sensitivity to evaluate models for screening clinical review titles and abstracts. Sensitivity is calculated by comparing model decisions to human reviewers.
\cite{suppadungsuk2023examining} evaluated ChatGPT's recall in identifying references for nephrology literature reviews. Recall was calculated as the percentage of existing references out of the total references generated, verified via PubMed, Google Scholar, and Web of Science.

For NLP tasks, recall is the most frequently used completeness metric. 
\cite{jahan2024comprehensive} uses recall to evaluate LLMs in NLP tasks, including NER, relation extraction, and entity linking. Recall measures the proportion of true positives among all actual positives.
\cite{alsentzer2023zero} evaluated the Flan-T5 model for extracting postpartum hemorrhage (PPH) concepts using recall. 
\cite{lituiev2023automatic} used recall for extracting social determinants of health (SDoH) in NER tasks from clinical notes of chronic low back pain (cLBP) patients.
\cite{hendrix2023trends} used recall (sensitivity) to evaluate an NLP algorithm for detecting pulmonary nodules in radiology reports.
\cite{morena2023clinical} used recall to evaluate completeness in identifying idiopathic pulmonary fibrosis cases from electronic health records.

In QA applications, assessing completeness reflects the quality of LLMs' responses, with higher completeness levels enhancing the efficiency of the question-answering process. 
\cite{pushpanathan2023popular} evaluated the comprehensiveness of LLM responses to ocular symptom inquiries using a five-point scale (1: not comprehensive to 5: very comprehensive), based on expert consensus ratings.
\cite{singhal2023large} used completeness metrics to evaluate LLMs on medical question-answering tasks by assessing the omission of important content in answers, with ratings based on clinical significance determined by clinician evaluations.
\cite{lim2023benchmarking} assessed the comprehensiveness of ChatGPT-3.5, ChatGPT-4.0, and Google Bard in answering myopia-related questions using a five-point scale, with scores based on detail level.
\cite{goodman2023accuracy} evaluated completeness using a 3-point Likert scale: 1 for incomplete, 2 for adequate, and 3 for comprehensive. This metric assessed chatbot-generated responses to physician-developed medical queries.
The study\cite{zack2024assessing} assessed GPT-4's completeness by examining detail levels in medical education, diagnostic reasoning, clinical plan generation, and patient assessment outputs, using expert comparisons and statistical significance tests for differential diagnosis detail accuracy.

In clinical application, including diagnosis and clinical predicting, recall is used to reflect the completeness of output of LLMs. 
In \cite{mermin2023use}, the NLP model classified patient-initiated EHR messages for COVID-19 case identification and treatment, and authors used recall as a completeness metric.
\cite{jiang2023health} used recall to evaluate NYUTron’s performance. It was applied to tasks like 30-day readmission prediction and in-hospital mortality prediction.

\textbf{Composite Metrics: }

Some evaluation metrics are designed as hybrids, combining multiple aspects into a consolidated set of composite indicators. This approach simplifies the evaluation process, allowing researchers to efficiently assess LLM performance by focusing on a few key metrics. Typically, these metrics are categorized into explicit and implicit types: explicit metrics involve a weighted combination of several indicators, while implicit metrics consider multiple factors concurrently during the evaluation process.

\begin{itemize}

\item \textbf{Explicit Composite Metrics: }
The F-score, particularly the F1-score, is one of the most commonly used explicit composite evaluation metrics, combining precision and recall to assess both correctness and completeness. The F1-score is calculated as the harmonic mean of precision and recall, effectively balancing the two aspects. This metric is frequently utilized in NLP tasks such as text classification, entity recognition, and relation extraction\cite{peng2023study, jahan2024comprehensive, alsentzer2023zero, lituiev2023automatic, buonocore2023localizing, morena2023clinical, nowak2023transformer, chen2023extensive}. For instance, it's applied in detecting fake news\cite{alghamdi2023towards}, including and excluding scientific literature\cite{guo2024automated}, and categorizing images\cite{sun2023ai}. In medical applications, the F1-score is crucial for tasks like diagnosis, clinical prediction, risk assessment, and prognosis prediction, demonstrating its versatility across various tasks and settings\cite{mermin2023use, jiang2023health, liu2023attention, sun2023ai}.

Besides, the BLURB score introduced in \cite{chen2023extensive} is a composite evaluation metric designed as part of the BLURB benchmark for biomedical language models. It represents a macro average of scores across multiple NLP tasks within the biomedical field. By aggregating performance across various tasks, the BLURB score provides a holistic view of a model's overall capabilities in handling complex biomedical text processing challenges, making it a comprehensive measure for assessing the effectiveness of language models in the biomedical domain.

Additionally, the AUC (Area Under the Curve) and AUROC (Area Under the Receiver Operating Characteristic curve) are frequently used as composite explicit evaluation metrics\cite{nowak2023transformer, jiang2023health, liu2023medical, liu2023attention}. They are for assessing the performance of binary classification models across all possible classification thresholds. Commonly applied in medical diagnostics and predictive modeling, they are essential for tasks where sensitivity and specificity are crucial, such as disease screening and patient risk assessment.

\item \textbf{Implicit Composite Metrics: }

In \cite{sun2023ai}, a composite metric was used to evaluate ChatGPT's responses to common questions about the ketogenic diet for diabetes management. This metric involved expert evaluations of answer quality based on professionalism, logical coherence, readability, and accuracy. Responses were rated as "unacceptable," "acceptable," or "excellent" by clinical nutrition experts, ensuring comprehensive assessment of the model's proficiency in providing medical nutrition therapy.
\cite{sezgin2023clinical} evaluated the quality of responses from GPT-4 (ChatGPT) and LaMDA (Bard) for postpartum depression (PPD) FAQs using a composite metric based on clinical accuracy and completeness. The assessment involved comparing LLM responses to American College of Obstetricians and Gynecologists (ACOG) standards, with ratings informed by the GRADE scale, which assesses the quality of evidence and strength of recommendations. 
\cite{mago2023potential} utilized a composite metric combining accuracy and completeness to evaluate the performance of ChatGPT-3 in oral and maxillofacial radiology report writing with a 4-point Likert scale. This assessment encompassed identifying radiographic anatomical landmarks, understanding oral and maxillofacial pathologies, and describing their radiographic features.
\cite{dhanvijay2023performance} utilized the Structural of Observed Learning Outcome (SOLO) taxonomy as a composite metric, evaluating accuracy and relevance to assess the performance of LLMs in answering physiology case vignettes. The SOLO metric rated responses on a scale from 0 to 4, covering aspects from pre-structural to extended-abstract levels, ensuring a comprehensive assessment of LLM-generated answers in the context of medical education.
\cite{wilhelm2023large} used the mDISCERN score, a composite metric evaluating the quality of medical information generated by LLMs. The mDISCERN score covers clarity of treatment options, objectives, balance, shared decision-making, mode of action, benefits, quality of life impact, risks, and additional sources. This evaluation was applied to therapeutic recommendations for 60 diseases in ophthalmology, dermatology, and orthopedics.
The study\cite{golan2023chatgpt} used the DISCERN tool, a composite metric evaluating the quality of medical information, covering aspects such as clarity, relevance, and bias, to assess ChatGPT's performance in evaluating online content about shock wave therapy for erectile dysfunction. 
Plus, medical examination scoring can also be seen as an implicit composite evaluation metric, as it is designed to assess the proficiency of medical professionals from multiple perspectives. For instance, the scoring system in \cite{li2023chatgpt} considers various aspects of the examination, including safety, communication, information gathering, and the application of clinical knowledge, effectively measuring comprehensive competencies in medical practice.

\end{itemize}

\textbf{Usability: }

When considering the comprehensive application of LLMs in the medical field, usability metrics such as helpfulness, safety, human-likeness, and robustness are crucial. These indicators assess the real-world effectiveness and reliability of LLM outputs, addressing not only technical performance but also user experience and ethical responsibility. Metrics like helpfulness and human likeness evaluate the model's ability to understand user intents and deliver practical information, while robustness and safety assessments explore consistency across demographics, potential biases, and harmful outcomes. These usability aspects are key to determining the success of LLM applications in healthcare settings.

\begin{itemize}

\item \textbf{Helpfulness \& Human-likeness: }
LLMs are good at understanding human inquires and generating helpfulness content.
\cite{singhal2023large} used relevance to user intent and helpfulness metrics, evaluated by lay users, to assess LLMs on consumer medical question-answering tasks. These metrics measured how well answers addressed users' questions and their overall helpfulness.
For better evaluating how well the content generated by LLMs, metrics for evaluating readability or understandability are needed. \cite{liu2023utility, cadamuro2023potentials, golan2023chatgpt, peng2023study} had demonstrated related metrics. 
\cite{liu2023utility} utilized reliability and humanness metrics to evaluate LLMs in clinical applications. Reliability included consistency (Cronbach's alpha), relevance, and readability/understandability. Humanness was assessed through the humanization of clinical letters, scoring the writing style's naturalness. These metrics were applied to generating clinical letters, radiology reports, and medical notes, demonstrating the model's effectiveness in producing human-like, reliable, and understandable medical documentation.
\cite{cadamuro2023potentials} evaluated ChatGPT's helpfulness in interpreting laboratory test results using an ordinal scale from 1 (very low) to 6 (very high). Helpfulness encompassed relevance, correctness, and the ability to provide non-obvious insights, appropriate suggestions, and enhanced patient comprehension. This metric aimed to assess the utility of ChatGPT's interpretations for laypersons in understanding medical content and making informed healthcare decisions.
The study\cite{golan2023chatgpt} employed seven readability indices (Flesch-Kincaid Level, Gunning-Fox Index, Coleman-Liau Index, SMOG Index, Automated Readability Index, FORCAST Grade Level, Flesch Reading Ease) to evaluate ChatGPT's readability assessment capabilities on online content about shock wave therapy for erectile dysfunction.
In \cite{peng2023study}, linguistic readability was used to evaluate GatorTronGPT's helpfulness and human-likeness in generating clinical text, considering clarity and coherence on a 1 (worst) to 9 (best) scale. 

\cite{peng2023study} also used a Turing test to evaluate GatorTronGPT's human-likeness by determining if physicians could distinguish between AI-generated and human-written clinical text, assessing the ability to mimic human writing. Similarly, \cite{majovsky2023artificial} and \cite{bernstein2023comparison} also evaluated the differences from LLMs and human. \cite{majovsky2023artificial} utilized helpfulness and human-likeness metrics, including patterns and probabilities in word choices and AI text classifiers, to evaluate the capability of AI language models, specifically ChatGPT, in generating high-quality fraudulent scientific articles. 
\cite{bernstein2023comparison} used the identification of response source as a metric to evaluate the human-likeness of ChatGPT-generated ophthalmology advice. Expert ophthalmologists assessed whether responses were written by AI or humans based on a 4-point scale.

\item \textbf{Robustness: }
Robustness is an evaluation metric that measures the performance consistency of a model under various influencing factors, which is essential when deploying models in real-world scenarios. The complexity and diversity of the real world introduce numerous variables, making robustness checks crucial for reliable applications. 
\cite{rao2023assessing} evaluated robustness metrics, specifically the impact of demographics (age and gender) and clinical acuity (Emergency Severity Index) on ChatGPT's performance in clinical decision support. These metrics assessed ChatGPT's accuracy in iterative clinical reasoning, including initial workup, diagnosis, and management, across different patient profiles and case severities.
\cite{kuroiwa2023potential} used the Fleiss $\kappa$ coefficient to evaluate the precision of ChatGPT in diagnosing five common orthopedic conditions. This metric assessed the consistency of answers provided by ChatGPT across different days and evaluators. Results indicated variability in reproducibility, ranging from poor to almost perfect, highlighting the influence of time and evaluators on response consistency.
\cite{zack2024assessing} used robustness metrics to evaluate outcome content bias in GPT-4’s diagnostic suggestions and patient assessments across racial and gender demographics. It examined differences in diagnosis rank, treatment recommendations, and subjective patient perceptions in clinical applications.
\cite{singhal2023large} used bias evaluation metrics to assess LLMs on consumer medical question-answering tasks, focusing on whether answers contained information inapplicable or inaccurate for specific demographics, evaluated by clinician reviewers.

\item \textbf{Safety: }
Safety is another crucial consideration, especially for medical applications involving direct patient interaction, such as clinical use. It is imperative to carefully and rigorously evaluate the performance of LLMs in terms of safety to ensure they do not pose risks to patients.
\cite{bernstein2023comparison} evaluated the safety of ChatGPT-generated ophthalmology advice using metrics for likelihood and extent of harm. Expert ophthalmologists assessed responses for potential harm and severity, comparing AI-generated and human-written answers to patient questions from an online forum.
In \cite{wilhelm2023large}, the "Potential Harmfulness" safety metric evaluates the risk associated with AI-generated medical advice. It assesses whether the content could lead to harmful outcomes if followed. This metric is applied to evaluate the safety of LLMs used for generating therapeutic recommendations in medical applications.
\cite{cadamuro2023potentials} evaluated ChatGPT's safety in interpreting laboratory test results using an ordinal scale from 1 (very low) to 6 (very high). Safety assessed the potential negative consequences of ChatGPT's responses on patient health, considering any harmful information or inadequate recommendations provided.
\cite{singhal2023large} used potential harm metrics to assess LLMs on consumer medical question-answering tasks, focusing on the severity and likelihood of health-related harms in answers, as evaluated by clinicians.
\cite{fraser2023comparison} used the unsafe triage rate as a safety metric to evaluate the clinical application of LLMs and SCs. The unsafe triage rate assessed how often recommendations from ChatGPT, WebMD, and Ada Health SCs potentially put patients at risk by underestimating the urgency of their conditions.

\item \textbf{Self-correction: }

The ability of LLMs to self-correct when prompted also demonstrates their capability. However, many studies did not consider the self-correction potential of LLMs, and only a few research efforts evaluate their capacity for self-correction.
\cite{pushpanathan2023popular} evaluated the self-checking ability of LLMs in ocular symptom inquiries by prompting the models to verify their responses. The assessment considered whether the models revised their answers or acknowledged inaccuracies when asked, "Could you kindly check if your answer is correct?"
\cite{brin2023comparing} used a self-correction metric to evaluate the consistency and confidence of ChatGPT and GPT-4 in answering USMLE-style soft skills questions. This involved a follow-up query, "Are you sure?", to assess if the models revised their initial responses. 
\cite{lim2023benchmarking} evaluated the self-correction capability of ChatGPT-3.5, ChatGPT-4.0, and Google Bard in responding to myopia-related questions by prompting them to review and correct initially rated 'poor' responses, then reassessing accuracy.

\item \textbf{Others: }

\cite{zack2024assessing} used diversity metrics to assess the innovativeness of GPT-4’s diagnostic and treatment plans. It evaluated GPT-4’s ability to generate varied and comprehensive differential diagnoses and treatment recommendations in clinical applications.
\cite{kuroiwa2023potential} evaluated ChatGPT's recommendations for medical consultation using key phrases such as “essential,” “recommended,” “best,” and “important.” This assessed the strength and frequency of recommendations to seek medical attention for self-diagnosed orthopedic conditions.
\cite{chervenak2023promise} used sentiment polarity as an evaluation metric to assess the emotional tone of responses to infertility FAQs. It also employed subjectivity scores to evaluate the degree of objectivity in responses to infertility FAQs. Both of the metrics are compared between ChatGPT's and CDC's answers.
\cite{fraser2023comparison} used the "too cautious" triage rate as a metric to evaluate LLMs and SCs in clinical applications. This metric assessed how often recommendations were overly cautious, potentially leading to unnecessary urgent care, with Ada and WebMD showing moderate rates.

\end{itemize}

\textbf{Consistency and Similarity Metrics: }

In addition to usability metrics, consistency and similarity metrics are crucial for assessing the alignment between LLM outputs and human medical consensus or professional outputs. Models that perform well in these metrics produce high-quality outputs that mirror expert opinions, thereby enhancing their reliability and user experience in practical applications. This part explores various metrics used to evaluate the agreement and similarity between LLM outputs and established medical standards or human expert responses.

\begin{itemize}

\item \textbf{Consensus: }

In \cite{sun2023ai}, the consistency rate with public consensus was used to evaluate ChatGPT's dietary recommendations for a ketogenic diet in diabetes management. This metric assessed how well the AI's recommendations aligned with expert guidelines from the China Low Carbon Diet Association, covering recommended and non-recommended foods. 
\cite{bernstein2023comparison} assessed the alignment of ChatGPT-generated ophthalmology advice with medical community consensus. Expert ophthalmologists evaluated whether the responses conformed to or opposed perceived medical standards, comparing AI-generated answers to human-written responses from an online forum. 
\cite{singhal2023large} used alignment with scientific consensus metrics to assess LLMs on consumer medical question-answering tasks. Clinicians evaluated whether the model's answers were aligned with prevailing scientific and clinical guidelines, opposed to consensus, or if no consensus existed.

\cite{guo2024automated} used Cohen's kappa to evaluate the agreement between GPT models and human reviewers in screening clinical review titles and abstracts. It considers the consistency of decisions related to clinical, decision-making, and medical consensus. The inter-rater variability between human screeners and agreement between model and human decisions were both evaluated. The inter-rater consistency not only can be used in the scenario of agreement between evaluator and annotator, but also can be used between model and human. 
For example, \cite{kumari2023large} used the Intraclass Correlation Coefficient (ICC) to evaluate inter-rater consistency for LLM-generated responses in hematology question-answering tasks, ensuring reliable assessment of accuracy by multiple raters. And \cite{lituiev2023automatic} evaluated inter-rater agreement using Cohen’s Kappa and F1 Score to assess annotator consistency for extracting social determinants of health (SDoH) in NER tasks from clinical notes of chronic low back pain (cLBP) patients. Those related metrics can be considered being used in evaluation of the consistency between content generated between AI and human.  

\item \textbf{Similarity: }
Text similarity metrics evaluate the degree of resemblance between texts from different sources, often used to assess how closely LLM-generated content aligns with human expert outputs. For instance:
\textbf{BLEU Score: }Measures the precision of n-grams between the machine-generated text and a set of reference texts, emphasizing the importance of exact matches.
\textbf{BERT Score: }Utilizes the BERT language model to compare the contextual embedding of words, providing a more nuanced measure of semantic similarity.
\textbf{METEOR: }Enhances evaluation by aligning and scoring based on exact, stem, synonym, and paraphrase matches between texts, offering a balanced approach to assessing meaning.
\textbf{ROUGE: }Focuses on recall by counting overlapping n-grams and word sequences, useful for evaluating summarization tasks.
\textbf{CIDEr: }Weighs human consensus by comparing n-gram frequencies in generated text against reference texts, ideal for capturing the salience of content.
\cite{nicolson2023improving} employs BLEU, METEOR, ROUGE-L, and CIDEr to evaluate the generated CXR reports' content consistency and similarity to radiologists' reports. These metrics consider n-gram overlap, precision, recall, and term frequency-inverse document frequency (TF-IDF) for comprehensive evaluation.
In \cite{liu2023medical}, BLEU, ROUGE, and CIDEr are used to assess the generated report's content consistency and relevance. They evaluate the effectiveness of the Med-MLLM in automatically generating coherent medical reports from radiology images.
\cite{tang2023evaluating} evaluated the use of LLMs for medical evidence summarization using ROUGE-L, METEOR, and BLEU metrics, covering recall, precision, and n-gram overlap between generated and reference summaries. 
\cite{jahan2024comprehensive} uses ROUGE and BERTScore to evaluate LLMs in text summarization. ROUGE measures lexical similarity, while BERTScore assesses semantic similarity between generated and reference summaries.

In addition to those frequently used metrics mentioned above, some other metrics that are not commonly seen can also be taken into consideration, including statistics methods, edit distance-based metrics and human manually evaluation. 
\cite{chen2023extensive, jin2023medcpt} used Pearson correlation coefficient as a metric for evaluating sentence similarity tasks, assessing the alignment between predicted and true similarity scores in biomedical and clinical sentence retrieval applications.
\cite{levkovich2023suicide} evaluated the performance of ChatGPT-3.5 and ChatGPT-4 in assessing suicide risk using Z-scores to measure discrepancies between AI and mental health professionals’ assessments, focusing on psychache, suicidal ideation, risk of suicide attempts, and resilience.
\cite{cadamuro2023potentials} evaluated ChatGPT's relevance in interpreting laboratory test results using an ordinal scale from 1 (very low) to 6 (very high). Relevance measured the coherence and consistency between ChatGPT's interpretation and the provided test results.
\cite{de2023open} uses edit distance-based similarity metrics, including Segmentation Similarity (S), Boundary Similarity (B), and the improved B2, to evaluate large language models' ability in paragraph segmentation by assessing the similarity between paragraphs.

\end{itemize}

\subsection{Benchmarks \& Datasets}

\subsubsection{General Benchmarks}

Some general benchmarks provide a wide range of datasets for testing LLMs across different tasks in medical domain. Usually they provide a series of datasets for different tasks. They can be seen as tools to evaluate LLM capabilities, which facilitate improvements and refinements based on the empirical results. 

BLURB (Biomedical Language Understanding \& Reasoning Benchmark)\cite{gu2021domain}is a comprehensive benchmark for evaluating LLMs. It is designed for NLP tasks such as NER, relation extraction, text classification, and question answering. It utilizes a diverse set of datasets including BC5CDR-CHEM, BC5CDR-Disease\cite{li2016biocreative}, BC2GM\cite{smith2008overview}, JNLPBA\cite{collier2004introduction}, EBM PICO\cite{nye2018corpus}, CHEMPROT\cite{taboureau2010chemprot}, GAD\cite{bravo2015extraction}, BIOSSES\cite{souganciouglu2017biosses}, HoC\cite{baker2016automatic}, PubMedQA\cite{jin2019pubmedqa}, and the BioAsq challenge\cite{tsatsaronis2015overview}.

\cite{singhal2023large}introduces MultiMedQA, a benchmark for evaluating LLMs in the medical domain. The benchmark encompasses various datasets including PubMedQA\cite{jin2019pubmedqa}, MedQA\cite{jin2021disease}, MedMCQA\cite{pal2022medmcqa}, LiveQA\cite{abacha2017overview}, MedicationQA\cite{Riveros2005StudiesIH}, MMLU clinical topics\cite{hendrycks2020measuring}. Moreover, it introduces a new dataset, HealthSearchQA, which covers medical questions spanning from professional exams to consumer health inquiries.

The CBLUE(Chinese Biomedical Language Understanding Evaluation)\cite{zhang2021cblue} is a general benchmark in Chinese. It encompasses several datasets, such as CMeEE, CMeIE\cite{guan2020cmeie}, CHIP-CDN, CHIP-STS, CHIP-CTC, KUAKE-QIC, KUAKE-QTR, and KUAKE-QQR, designed to cover tasks from named entity recognition to query-document relevance.

Another Chinese general benchmark, MedBench\cite{cai2024medbench} which is a comprehensive benchmark as well. It includes datasets from the Chinese Medical Licensing Exam, Resident Standardization Training Exam, Doctor In-Charge Qualification Exam, and real-world clinical cases, which totally includes 40,041 questions across various branches of medicine. There are many datasets on its official website ranging from medical question answering, medical language understanding, complex medical inference etc\cite{liu2022meddg, chen2023benchmark, wang2023cmb, hongying2021building, guan2020cmeie, zhu2023overview, zong2021semantic}. 

\subsubsection{Benchmarks for Question Answering}

Following the general benchmarks discussed above, this section shifts focus to benchmarks specifically designed for Question Answering (QA) tasks. These benchmarks are categorized based on their relevance to specific QA scenarios: clinical decision support, involving medical dialogue and consultation cases; medical knowledge QA, which includes questions from medical exams and literature-based QAs; and medical information retrieval includes datasets from publicly accessible medical records, consumer health inquiries, and public medical websites. 

\textbf{Clinical Decision Support: }

The MedDG dataset detailed in \cite{liu2022meddg} includes 17,864 Chinese medical dialogues, with 385,951 utterances and 217,205 entities, originating from the online medical consultation platform Doctor Chunyu. It is designed for entity extraction initially, but the dataset can be adapted for training LLMs on clinical consultation tasks. The original evaluations precision, recall, and F1 scores are for NER tasks, adapting this dataset for dialogue generation might use other metrics such as BLEU and Rouge-L, which were not discussed in the original paper but are usually for assessing response quality in conversational models. 

The CMB-Clin dataset orginated from \cite{wang2023cmb}, is designed for clinical diagnostic scenarios. It consists of 74 expert-curated medical consultations derived from real case studies and totally 208 questions. These datasets are from official medical textbooks and they are refined through  quality screening process to exclude similar diseases to ensure diversity. The primary benchmarking task for CMB-Clin involves a multi-turn dialogue-based question-answering format, which evaluates the medical knowledge application of LLMs. Evaluation metrics include Fluency, Relevance, Completeness, and Proficiency, and it is assessed both through expert and GPT-4, which ensures robust performance measurement.

The LongHealth benchmark\cite{adams2024longhealth} includes 20 fictional patient cases, each of them is in real-world discharge note structures. These cases are utilized to evaluate LLMs on tasks including information extraction, negation, and time variant information extraction. Models are tested with 400 multiple-choice questions, with performance measured with accuracy in extracting or identifying correct or missing information.

\textbf{Medical Knowledge QA: }

The Professional Medicine task in \cite{hendrycks2020measuring} employs questions from the US Medical Licensing Examination. It consists of 15,908 questions, and it is designed to test real-world medical knowledge. 
The MedQA dataset\cite{jin2021disease}, derived from medical board exams in the US, Mainland China, and Taiwan, comprises 61,097 questions across three languages. It serves as a benchmark for open-domain question answering tasks. 
MedMCQA\cite{pal2022medmcqa} is a large-scale multiple choice QA dataset derived from AIIMS \& NEET PG entrance exams, featuring over 194,000 questions across 21 medical subjects and 2.4k healthcare topics. 
The CMB-Exam\cite{wang2023cmb} is another multiple choice QA dataset as well. It is derived from publicly available medical qualification exams in China, encompasses 280,839 multiple-choice questions across various medical professions. 
The datasets from medical examinations above is used for benchmarking the knowledge comprehension of LLMs, and their evaluation is usually based on accuracy.

Apart from medical examination questions, \cite{jin2019pubmedqa} introduced PubMedQA, a biomedical QA dataset, features 1,000 expert-annotated, 61,200 unlabeled, and 211,300 artificially generated instances derived from PubMed abstracts of scientific researches. It is in the form of yes/no/maybe question answering. Evaluation metrics include accuracy and macro-F1, mainly for evaluating the ability of reasoning over biomedical texts, especially quantitative content.

\textbf{Medical Information Retrieval QA: }

MEDIQA 2019\cite{abacha2019overview} provided datasets for three tasks: Natural Language Inference (NLI), Recognizing Question Entailment (RQE), and Question Answering (QA). The datasets is originated from clinical sentence pairs (MedNLI), consumer health questions, and FAQs from NIH institutes, then it can be used to build models with capabilities to help the public retrieve information in the form of QA. Evaluation metrics included accuracy, Mean Reciprocal Rank (MRR), and Spearman’s Rank Correlation Coefficient. 

The TREC 2017 LiveQA medical task\cite{abacha2017overview} utilized datasets in which there are 634 medical question-answer pairs, sourced from the U.S. National Library of Medicine. The datasets were used for accessing models' ability in answering consumer health questions. Evaluation metrics were based on average scores and precision.

\subsubsection{Benchmarks for Summarization}

In this part, we are going to discuss benchmarks for summarization. LLMs have been proven its capabilities of language understanding and abstraction. These models are good at extracting key information from texts across various contexts such as QA content, dialogues, clinical evidence, and medical reports. The categorization of benchmarks for summarization can be roughly split into five types ranging from questions and answers summarization to transforming professional medical knowledge into lay summaries and abstracting medical reports.

\textbf{Question Summarization: }

The MeQSum corpus\cite{abacha2019summarization} consists of 1,000 consumer health question summaries. It originates from a dataset distributed by the U.S. National Library of Medicine. Its primary task is to evaluate abstractive summarization models, specifically focusing on generating condensed versions of consumer health questions. Evaluation metrics are primarily based on ROUGE scores.

MEDIQA 2021\cite{abacha2021overview} introduced three datasets for medical text summarization: the MeQSum dataset, MEDIQA-AnS, and radiology reports from Indiana University and Stanford Health Care. They served as benchmarks for summarizing consumer health questions, multiple answer aggregation, and radiology report impressions, respectively. Evaluation metrics included ROUGE-2, BERTScore, and HOLMS.

\cite{mrini2021gradually} introduced MeQSum, HealthCareMagic, iCliniq and Recognizing Question Entailment (RQE) . MeQSum, sourced from the U.S. NIH, includes 1,000 samples. HealthCareMagic and iCliniq, derived from online platforms, contain 226,405 and 31,062 entries respectively. RQE includes 9,120 questions, primarily focusing on matching consumer health questions to expert-answered FAQs. These datasets support benchmarking summarization tasks where the primary evaluation metric is the ROUGE score.

\textbf{Answer Summarization: }

The MEDIQA-AnS dataset\cite{savery2020question} comprises 156 consumer health questions, their corresponding answers, and expert-created summaries. It is from the MEDIQA-QA dataset, and MEDIQA-AnS extends it with both extractive and abstractive summaries. It supports evaluations through ROUGE and BLEU metrics. Besides, \cite{abacha2021overview} also provides the dataset for answer summarization.  


\textbf{Lay Summary: }

The corpus\cite{luo2022readability} for readability controllable summarization comprises 28,124 biomedical documents, sourced from peer-reviewed journals like PLOS Medicine. This dataset is designed to benchmark controllable abstractive and extractive summarization models which generate both technical for experts and plain language for layman. Evaluation metrics focus on readability levels, comparing plain language summaries to technical summaries using a novel masked language model-based metric to assess readability discrepancies.

As introduced in \cite{goldsack2022making} and \cite{goldsack2023overview}, the PLOS and eLife datasets from biomedical journals form the basis for lay summarization benchmarks. The datasets contain 24,773 articles from PLOS and 4,346 from eLife. PLOS is extensive with short author-written summaries, whereas eLife features longer, expert-edited summaries. Evaluation metrics include readability scores and expert-based manual assessments. They benchmark lay summarization models' ability to generate comprehensible summaries for non-experts. Evaluation metrics include ROUGE scores, Flesch-Kincaid and Dale-Chall readability scores, and BARTScore for factuality.

\textbf{Medical Report Generation \& Summarization: }

Medical report generation can be roughly seen as a summarization task based the medical evidence includes consultation content, medical reviews and even multimodal data.  
The IMCS-21 dataset\cite{chen2023benchmark} is used for the Medical Report Generation task. It originates from extensive Medical Consultation Records collected from a Chinese online health community called Muzhi, it encompasses 4,116 annotated samples. This dataset serves as a benchmark for generating medical reports, summarizing consultations based on dialogues. Evaluation metrics include ROUGE scores and Regex-based Diagnostic Accuracy, assessing the generated reports' quality and medical accuracy. DeltaNet is evaluated on the COVID-19 dataset\cite{wu2022deltanet}, consisting of 1,261 exams from 1,085 patients. It also uses IU-Xray\cite{demner2016preparing} and MIMIC-CXR\cite{johnson2019mimic} datasets, the largest being MIMIC-CXR with 377,110 images and 227,827 reports. It benchmarks for automatic medical report generation based on image and report data. Performance metrics include BLEU, CIDEr, and ROUGE-L scores. In addition to public datasets, Cochrane Reviews serve as a valuable resource for evaluating clinical evidence summarization. In \cite{tang2023evaluating}, they evaluated LLMs on the task of medical evidence summarization based on the cochrane reviews of six different clinical domains\cite{muhlbauer2021antipsychotics, wang2023endovascular, gross2015exercises, kamo2022repetitive}.

\subsubsection{Benchmarks for Information Extraction}

Our focus shifts to information extraction. Information extraction is an important part of traditional natural language processing tasks . Beyond understanding and generating coherent responses, LLMs excel at extracting key information from complex texts to form structured, standardized data. For better accessing their ability, such as named entity recognition, relation extraction, entity linking, text classification, and comprehensive information extraction like event extraction and PICO from medical research texts, benchmarks and datasets need to be built.

\textbf{Named Entity Recognition: }

Named Entity Recognition is one of the most widely seen NLP task. 
The NCBI disease corpus\cite{dougan2014ncbi}, containing 793 fully annotated PubMed abstracts with 6,892 disease mentions linked to 790 unique disease concepts, originates from PubMed . This dataset serves as a benchmark for NER tasks in biomedical text mining. 
\cite{hongying2021building} constructs a corpus for pediatric medical text information extraction, comprising 38,805 medical entries segmented into nine categories. It is designed for tasks such as word segmentation and NER in clinical pediatric texts. 
The JNLPBA shared task\cite{collier2004introduction} utilizes the GENIA version 3.02 corpus, derived from MEDLINE abstracts selected using MeSH terms . This dataset comprises 2,000 abstracts for training and 404 for testing. It serves as a benchmark for bio-entity recognition tasks. 
The BioCreative II Gene Mention Task\cite{smith2008overview} utilized a dataset comprising 20,000 sentences with approximately 44,500 GENE and ALTGENE annotations, sourced from previous BioCreative challenges . This dataset served as a benchmark for testing gene mention recognition systems. 
The CHEMDNER corpus\cite{krallinger2015chemdner} comprises 10,000 PubMed abstracts with 84,355 manually annotated chemical entity mentions. Originating from PubMed, this dataset serves as a benchmark for chemical NER tasks. 
The LINNAEUS system\cite{gerner2010linnaeus} utilizes a manually annotated corpus of full-text articles as a dataset for NER benchmarks. This dataset is derived from PubMed Central’s open access subset, comprising 100 full-text documents. And \cite{pafilis2013species} includes the Linnaeus-100 and Species-800 datasets. Species-800 comprises 800 abstracts across diverse taxonomic fields . These two researches serve as benchmarks for evaluating species name recognition performance.
The CMeEE dataset\cite{guan2020cmeie}, sourced from CHIP 2020\cite{zhang2021cblue}, contains 23,000 labeled samples for NER. Entities are categorized into nine classes, such as diseases and drugs . 
Generally, evaluation metrics for NER tasks are calculated using precision, recall, and F-measure to assess the performance.

\textbf{Relation Extraction: }

Typically, the relation extraction is often referred as the relation among entities.
\cite{herrero2013ddi} comprises 792 texts from the DrugBank database and 233 Medline abstracts, annotated with 18,502 pharmacological entities and 5,028 drug-drug interactions . It serves as a benchmark for evaluating abilities focused on identifying pharmacological substances and detecting DDIs. 
The dataset KD-DTI\cite{hou2022discovering} originates from DrugBank and the Therapeutic Target Database (TTD). It includes 14,256 documents, encompassing 139,810 sentences and 3,671,000 words. This dataset is employed as a benchmark for drug-target interaction (DTI) discovery tasks. 
The BC5CDR dataset\cite{li2016biocreative} comprises 1,500 PubMed articles, annotated with 4,409 chemicals, 5,818 diseases, and 3,116 chemical-disease relations. Originating from the collaborative CTD-Pfizer corpus and additional articles selected specifically for BioCreative V .
The ChemProt database\cite{taboureau2010chemprot} contains over 700,000 unique chemicals annotated for 30,578 proteins and more than 2 million chemical-protein interactions. These data are sourced from repositories such as ChEMBL, DrugBank, and PubChem. It is primarily used to benchmark tasks in chemical-protein relation extraction. 
\cite{bravo2015extraction} discusses the use of the EU-ADR corpus and a semi-automatically annotated corpus from the GAD(Genetic Association Database) database in relation extraction tasks . The EU-ADR corpus, utilized for benchmarking drug-disease, drug-target, and gene-disease associations, involves annotations derived from expert consensus. 
\cite{luo2022biored} discusses the BioRED dataset, which includes 600 PubMed abstracts, focusing on the extraction of biomedical entities and their relations. This dataset benchmarks NER and relation extraction tasks . It captures multiple entity types like genes, diseases, and chemicals, assessing the models using precision, recall, and F-scores.
\cite{guan2020cmeie} discusses the CMeIE dataset, sourced from medical textbooks and clinical practices, comprising 28,008 sentences and 85,282 triplets. It benchmarks joint entity and relation extraction tasks.

\textbf{Entity Linking: }

\cite{basaldella2020cometa} introduced the COMETA dataset, comprising 20,000 expert-annotated biomedical entity mentions from health discussions on Reddit, linked to SNOMED CT. This dataset is employed as a benchmark for Entity Linking tasks to assess how models link health-related terms to medical concepts . Evaluation metrics include top-k Accuracy and Mean Reciprocal Rank. Moreover, the NCBI disease corpus\cite{dougan2014ncbi}(mentioned in the part of NER) not only includes annotations for NER but also provides data for entity linking or normalization. Each disease mention is linked to a standard disease concept in either the Medical Subject Headings (MeSH) or Online Mendelian Inheritance in Man (OMIM) databases. The CHIP-CDN dataset\cite{zhang2021cblue} contains 18,192 samples sourced from Chinese electronic health records (EHRs) for clinical diagnosis normalization, and it is annotated based on the ICD-10 standard.

\textbf{Text Classification: }

\cite{zong2021semantic} includes 75,754 Chinese and English eligibility criteria sentences from the Chinese Clinical Trial Registry (ChiCTR). After preprocessing, 19,185 sentences were used for unsupervised clustering and 38,341 for supervised classification tasks. 
\cite{baker2016automatic} introduces a dataset comprising 1,499 PubMed abstracts annotated based on the evidence they provide for the 10 hallmarks of cancer. This dataset originates from PubMed and serves as a benchmark for testing text classification performance of models. 
The LitCovid dataset\cite{chen2021litcovid} consists of COVID-19-related articles retrieved from PubMed. As of August 2020, the dataset includes approximately 80,000 articles used for testing document classification tasks. 
\cite{zhang2021cblue} includes two datasets for classification task. One is CHIP-CTC containing 40,644 samples collected from the Chinese Clinical Trial Registry (ChiCTR), it is used for classifying clinical trial eligibility criteria into 44 categories. The other is KUAKE-QIC , containing 10,880 samples sourced from real-world search engine logs, it is used for intent classification into 11 medical intent categories. The evaluation metric for this dataset is accuracy.
The performance of classification models is often evaluated using Precision, Recall, micro F1-score, and accuracy.

\textbf{Comprehensive Information Extraction: }

Comprehensive information extraction task, including temporal reasoning, medical event extraction, and PICO extraction for RCT medical research, offers a more flexible approach compared to traditional NER or relation extraction. It enables a broader and more nuanced understanding of complex data structures in clinical narratives and research texts, enhancing the depth and applicability of extracted information.
For medical event extraction task, the dataset used in the PromptCBLUE shared task includes the CHIP-CDEE sub-task\cite{zhu2023overview}, which focuses on clinical finding event extraction . Originating from the CBLUE benchmark, the CHIP-CDEE dataset is designed to evaluate models on medical information extraction tasks. The dataset features a total of 3000 training samples and 400 each for validation and testing. 
The dataset\cite{sun2013annotating} includes 310 de-identified discharge summaries from Partners Healthcare and Beth Israel Deaconess Medical Center, comprising approximately 178,000 tokens. It serves as a benchmark for structured information extraction tasks in clinical narratives, focusing on temporal reasoning. Evaluation metrics involve precision and recall for EVENT and TIMEX3 annotations, and various methods to assess TLINK annotations for consistency and accuracy.
The EBM-NLP dataset\cite{nye2018corpus} comprises 5,000 annotated medical abstracts sourced from PubMed's MEDLINE database, specifically targeting clinical trials. These annotations focus on the PICO framework (Patient population, Interventions, Comparators, Outcomes) to support tasks like information extraction. Evaluation metrics are computed using aggregation strategies like majority voting and advanced models like Dawid-Skene and HMM-Crowd, measuring precision, recall, and F-1 scores.

\subsubsection{Benchmarks in Bioinformatics}

Although LLMs are traditionally seen as tools for natural language processing, their capabilities extend robustly into handling and reasoning with sequential data. For instance, in bioinformatics, LLMs like the GPT have been adapted for tasks such as de novo molecular design, where they generate novel compounds by manipulating sequence-like molecular representations\cite{wang2023cmolgpt}. Similarly, LLMs have been employed to analyze biological sequence similarities\cite{li2023bioseq} and infer RNA structure from homologous sequences\cite{zhang2024multiple}. These applications demonstrate LLMs' ability to extract and generate critical information from complex sequence data, underscoring their broader utility beyond conventional text processing.

\cite{polykovskiy2020molecular} uses the MOSES dataset derived from the ZINC Clean Leads collection, containing 4,591,276 molecules. These datasets are applied in the benchmarking of molecular generation tasks. Evaluation metrics include validity, uniqueness, novelty, and other molecular properties to assess the performance of generative models.

\cite{prykhodko2019novo} utilized the ChEMBL database and ExCAPE-DB, comprising 1,347,173 and targeted subset SMILES for GAN training. Benchmarks focused on generating novel, valid compounds. Evaluation metrics included validity, uniqueness, and novelty calculated by comparing generated and training set SMILES.

The ExCAPE-DB dataset\cite{sun2017excape} comprises 998,131 unique compounds and 70,850,163 SAR data points sourced from PubChem and ChEMBL. It is utilized as a benchmark for multi-target machine learning model evaluation, specifically assessing model performance using metrics such as sensitivity, precision, specificity, and Cohen's k.

The CAFA3\cite{zhou2019cafa} challenge utilized datasets from genome-wide mutation screenings in Candida albicans, Pseudomonas aeruginosa, and targeted gene assays in Drosophila melanogaster. These datasets comprised thousands of genes evaluated for functionalities like biofilm formation and motility. Performance metrics calculated included Fmax and Smin, based on precision-recall and semantic distance, respectively, incorporating information content weighting.

\cite{wei2021icircda} employs the CircFunBase dataset, containing 4195 verified circRNA-disease associations involving 3704 circRNAs and 90 diseases. This dataset serves as a benchmark for tasks like detecting missing associations and predicting new circRNA-related diseases. Performance metrics such as AUC, AUPR, MRR, and NDCG are used.

In the updated Rfam database, the datasets\cite{kalvari2021rfam} include multiple sequence alignments and covariance models from collaborations with ZWD, miRBase, and EVBC. The expanded Rfamseq now contains 14,772 genomes across all life domains. These datasets are primarily used for benchmarking non-coding RNA annotation in genomic sequences.

The MARS database\cite{chen2023master} integrates datasets from sources like NCBI, RNAcentral, MG-RAST, Genome Warehouse, and MGnify, totaling 1.73 billion sequences. It benchmarks RNA homology search tasks using RNAcmap3. Performance metrics are calculated using F1-scores, sensitivity, and precision from direct coupling analyses . The SCOP database\cite{murzin1995scop} provides a comprehensive catalog of protein structural and evolutionary relationships. It includes links to coordinates, structure images, and sequence data.  \cite{rose2016rcsb} discusses the RCSB Protein Data Bank, a global archive of atomic-level 3D structures of biological macromolecules sourced from laboratories worldwide.

\subsubsection{Benchmarks for Information Retrieval}

LLMs have shown significant text comprehension ability, which makes them potentially valuable tools for medical research by enabling literature discovery across multiple dimensions , for example, question retrieval, evidence retrieval, or fact-checking. Currently, benchmarks related to information retrieval typically include tasks like fact verification, citation prediction, and literature recommendation and so on . These benchmarks include various forms of similarity and relevance measures, which provides a evaluation framework for assessing LLMs' in medical information retrieval.

In the BEIR benchmark\cite{thakur2021beir}, five biomedical datasets are included: TREC-COVID, NFCorpus, BioASQ, SCIDOCS, and SciFact. These datasets, sourced from publicly available biomedical corpora, range from 3.6k to 15M documents. They are used to evaluate models in biomedical information retrieval tasks, such as question answering, citation prediction, and fact checking. Evaluation is based on nDCG@10 and other metrics.
The RELISH-DB dataset\cite{brown2019large} contains over 180,000 pairs of PubMed articles annotated by more than 1500 scientists from 84 countries. The annotations serve as benchmarks for evaluating document recommendation systems. Key metrics such as MCC, AUC, P@N, and MRR assess performance in binary classification and information retrieval tasks by comparing the relevance of recommended articles to seed articles.
SPECTER introduces SCIDOCS\cite{cohan2020specter}, a benchmark comprising datasets for information retrieval tasks, including citation prediction, document classification, and user activity prediction. The datasets contain over 23,000 medical papers for MeSH classification, alongside tens of thousands of examples derived from public scholarly search engine logs. Evaluation metrics include MAP and nDCG to measure ranking performance.
The BIOSSES dataset\cite{souganciouglu2017biosses} comprises 100 manually annotated sentence pairs sourced from the TAC Biomedical Summarization Track Training Dataset. Each pair is scored by five human experts from 0 (unrelated) to 4 (equivalent). It serves as a benchmark for evaluating sentence-level semantic similarity. The evaluation metric is the Pearson correlation between algorithm scores and human annotations.
The MedSTS dataset\cite{wang2020medsts}, sourced from Mayo Clinic's clinical corpus, contains 174,629 sentence pairs, with a subset, MedSTS\_ann, comprising 1,068 annotated pairs. It benchmarks semantic textual similarity (STS) systems for reducing redundant information in electronic health records. Evaluation uses Pearson correlation to compare system scores against human-annotated scores. 

CBLUE\cite{zhang2021cblue} introduces three datasets for measuring similarity or relevance on retrieval tasks.
The CHIP-STS dataset, containing 20,000 samples collected from Chinese disease question-and-answer data, is used for evaluating sentence similarity in a non-i.i.d. setting across five disease types. The performance is measured using the Macro F1 score.
For relevance measuring, there are two dataset, KUAKE-QTR and KUAKE-QQR.
The KUAKE-QTR dataset, containing 32,552 samples sourced from real-world search engine logs, is used to estimate query-document relevance. The task involves determining the relevance of a query to a given document title.
The KUAKE-QQR dataset, containing 18,196 samples sourced from real-world search engine logs, is used for query-query relevance estimation. This task determines the semantic similarity between two queries. Both of them are evaluated using accuracy.

\section{Challenges and Issues}

Evaluating the application of LLMs in the medical field presents several technical, ethical, and legal challenges that need to be addressed to ensure their safe and effective use. Unlike general applications, the medical field demands more strict and detailed evaluation frameworks. This section delves into these challenges and discusses strategies for improving evaluation methods and metrics to overcome these issues.

\subsection{Technical Challenges}
In the context of disease management and outpatient care, several technical challenges hinder the effective evaluation of LLMs. One major issue is the lack of data and sample diversity. Current studies often rely on single datasets that fail to represent the full spectrum of patients across different ages, genders, race, and geographic locations, which limits the generalization of the results. Additionally, many evaluations focus on short-term effects, with insufficient follow-up studies to assess the long-term performance and impact of LLMs.

Another significant challenge is the interpretability of LLMs. The medical field requires a deep understanding of the decision-making processes behind diagnoses and treatment recommendations, but many models, especially commercial ones like the GPT series, remain "black boxes" with opaque internal mechanisms and undisclosed training data. This lack of transparency limits a thorough understanding of their performance.

Furthermore, evaluations mostly focus on text data, neglecting the potential of LLMs to integrate and process multi-modal data such as images, audio, and sensor data. Comprehensive data integration is particularly important for diagnostic and treatment support, where a combination of different data types can significantly enhance decision-making.

\subsection{Ethical and Legal Challenges}
Protecting patient privacy is a great ethical challenge. The use of sensitive medical data raises significant concerns about data breaches and misuse. Ensuring strict data encryption and access control measures is important to protect patient information. Moreover, compliance with relevant privacy regulations and standards must be maintained.

Bias and fairness are another critical ethical issue. LLMs can exhibit biases that affect their performance across different patient groups. These biases might be caused because of the data used for training and the algorithms themselves, potentially leading to discriminatory outcomes. Developing and applying bias mitigation, regular bias audits, and ensuring diverse representation in training datasets are potential solution to issue.

From a legal perspective, regulatory compliance is a complex challenge due to the differing regulations across regions. Establishing global collaboration to harmonize regulatory standards for LLMs in healthcare is essential. Working closely with legal experts can help ensure compliance and create clear guidelines and frameworks for regulatory adherence.

\subsection{Strategies for Improvement}
To address these challenges, more holistic evaluation frameworks are needed. Current frameworks often fall short in comprehensively assessing LLMs in medical applications, overlooking critical aspects such as usability, robustness, and safety. Developing frameworks that encompass technical performance, ethical considerations, and legal compliance is essential. Integrating multi-dimensional evaluation metrics can provide a comprehensive assessment of LLMs.

Improving evaluation methods and metrics is also crucial. Existing methods and metrics primarily focus on correctness and completeness, with less emphasis on usability, robustness, and safety. Introducing new metrics to evaluate the coherence, logic, and safety of generated content, combining qualitative and quantitative evaluation methods, and using advanced automated tools and human expert reviews can enhance evaluation accuracy and depth.

Finally, addressing gaps and limitations in current methodologies is necessary for advancing the field. Conducting systematic reviews and empirical studies to identify weak points in current evaluations, fostering collaboration between academia, industry, and healthcare practitioners to develop innovative solutions, and regularly updating and refining evaluation standards to keep pace with technological advancements and emerging challenges are vital steps to ensure the continuous improvement of evaluation practices.

\section{Conclusion \& Outlook}

This comprehensive survey highlights the potential and challenges associated with the application and evaluation of LLMs in the medical field. Through an in-depth analysis, we have underscored the necessity for specialized evaluation frameworks tailored to the unique demands of healthcare applications.

Our survey organizes the multiple type of roles of LLMs in clinical settings, medical text data processing, research, education, and public health awareness. By examining the various evaluation methods, including models, evaluators, and comparative experiments, we provide a detailed understanding of the metrics used to assess LLMs' effectiveness, accuracy, usability, and ethical alignment.

Key technical challenges include data quality and diversity, model interpretability, and the integration of multi-modal data. Addressing these requires developing more robust data collection methods, enhancing model transparency, and expanding evaluations to include diverse data types. Ethical and legal challenges, such as patient privacy, bias and fairness, and regulatory compliance, necessitate stringent data protection measures, bias mitigation strategies, and harmonized global regulations.

To improve the evaluation of LLMs in healthcare, we recommend developing holistic frameworks that encompass technical performance, ethical considerations, and legal compliance. Enhancing evaluation methods and metrics to include usability, robustness, and safety is of importance. Additionally, systematic reviews and collaborative efforts are needed to address current gaps and limitations, ensuring continuous improvement in evaluation practices.

Looking ahead, the responsible development and deployment of LLMs in healthcare will depend on ongoing empirical validation and the establishment of rigorous, multi-dimensional evaluation frameworks. By addressing the outlined challenges and implementing the suggested improvements, we can harness the full potential of LLMs to enhance healthcare outcomes while maintaining strict ethical standards. This survey aims to equip healthcare professionals, researchers, and policymakers with the insights needed to integrate and evaluate these powerful models effectively, ultimately contributing to safer, more effective, and ethically medical practices.

\section{Funding Statement}
This research was supported by the National Key R\&D Program of China (Grant No. 2022YFC2010105) and the Shenzhen Higher Education Stability Support Program (Project: Research on a Smart Analysis Model for Food Nutrition Based on Large Models).

\bibliographystyle{unsrt}  
\bibliography{references}

\end{document}